\documentclass{article}

\PassOptionsToPackage{numbers}{natbib}
\usepackage[eandd,final]{neurips_2026}

\usepackage[utf8]{inputenc}
\usepackage[T1]{fontenc}
\usepackage{graphicx}
\usepackage{tabularx}
\usepackage{booktabs}
\usepackage{amsmath}
\usepackage{amssymb}
\usepackage{amsfonts}
\usepackage{nicefrac}
\usepackage{microtype}
\usepackage[table]{xcolor}
\usepackage{colortbl}
\usepackage{caption}
\usepackage{subcaption}
\usepackage{xspace}
\usepackage{multirow}
\usepackage{longtable}
\usepackage{wrapfig}
\usepackage{url}
\usepackage[colorlinks=true,linkcolor=blue,citecolor=blue,urlcolor=blue]{hyperref}
\usepackage{tcolorbox}
\tcbuselibrary{listings,breakable,skins}
\usepackage{listings}

\definecolor{promptframe}{HTML}{4A6FA5}
\definecolor{promptback}{HTML}{F7F9FC}
\newtcolorbox{promptbox}[1]{
  colframe=promptframe,
  colback=promptback,
  coltitle=white,
  fonttitle=\bfseries\small,
  title=#1,
  breakable,
  enhanced,
  left=4pt,
  right=4pt,
  top=2pt,
  bottom=2pt,
  boxrule=0.6pt,
}

\definecolor{newcat}{HTML}{E8F5E9}
\definecolor{viscat}{HTML}{E3F2FD}
\definecolor{multicat}{HTML}{FFF3E0}

\newcommand{\modellogo}[1]{%
  \raisebox{-0.18ex}{\includegraphics[height=0.95em]{logos/#1}}\,%
}
\newcommand{\openailogo}{\modellogo{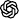}}
\newcommand{\anthropiclogo}{\modellogo{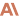}}
\newcommand{\googlelogo}{\modellogo{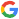}}
\newcommand{\metalogo}{\modellogo{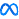}}
\newcommand{\xailogo}{\modellogo{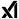}}
\newcommand{\deepseeklogo}{\modellogo{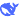}}
\newcommand{\qwenlogo}{\modellogo{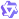}}
\newcommand{\zailogo}{\modellogo{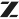}}
\newcommand{\modelname}[1]{\textbf{\texttt{#1}}}

\definecolor{ctext}{HTML}{249BEF}
\definecolor{cimage}{HTML}{F59E0B}
\definecolor{cvideo}{HTML}{DC2654}

\newcommand{\ours}{\textsc{Who\&When Pro}\xspace}

\title{\textsc{Who\&When Pro}: Can LLMs Really Attribute Failures in AI Agents?}

\author{%
  Jiale Liu$^{1,5}$ \quad
  Huajun Xi$^{2}$ \quad
  Shaokun Zhang$^{1}$\thanks{Corresponding author: \texttt{shaokun.zhang@psu.edu}} \quad
  Yifan Zeng$^{3}$ \quad
  Tianwei Yue$^{4}$ \\
  \textbf{Chi Wang$^{5}$ \quad
  Jian Kang$^{2}$ \quad
  Qingyun Wu$^{1,5}$ \quad
  Huazheng Wang$^{3,5}$} \\[0.5em]
  $^{1}$Penn State University \quad
  $^{2}$Mohamed bin Zayed University of Artificial Intelligence \\
  $^{3}$Oregon State University \quad
  $^{4}$Mathos AI \quad
  $^{5}$AG2ai, Inc.
}

\begin{document}

\maketitle


\begin{abstract}
Automated failure attribution~\citep{zhang2025which} uses LLMs to identify where and why agentic systems fail.
As agents become more capable, their failures become subtler, making automated attribution increasingly important.
We introduce \textsc{Who\&When Pro}, a large-scale benchmark for automated failure attribution in agentic systems.
Using a strictly controlled pipeline that injects a failure only after exactly replaying a successful prefix, we construct 12{,}326 failed trajectories with golden labels across 3 modalities and 26 benchmarks covering various scenarios.
Beyond benchmarking, we conduct extensive experiments and analyses, revealing systematic patterns in how models attribute failures across modalities, protocols, and model families, and providing empirical guidance for future automated failure attribution systems.
\end{abstract}

\section{Introduction}

Recent works show that model scaling~\citep{openai2026gpt54thinking, anthropic2026claudesonnet46systemcard, googledeepmind2025gemini3flashmodelcard}, together with careful harness engineering~\citep{lou2026autoharness, lin2026agentic, pan2026natural,lee2026metaharnessendtoendoptimizationmodel}, can produce AI agents capable of operating in diverse digital environments, with demonstrated potential across domains such as coding~\citep{gao2025trae, yang2024sweagent,xia2024agentless}, scientific discovery~\citep{ren2025towards,bran2024chemcrow, boiko2023coscientist}, and complex real-world problem solving~\citep{mialon2023gaia, liu-etal-2025-divide, zhang2025nemotron,liu2026teamfusion,song2024adaptive}.

However, advancing agentic systems becomes increasingly challenging as their capabilities grow: failures in more capable agents are more subtle and harder to detect, yet these failures provide crucial signals for further improvement.
This motivates \emph{automated failure attribution}, a research direction that uses LLMs to identify where and why agentic systems fail and convert failures into actionable feedback~\citep{zhang2025which,cemri2025why}. 
The goal is to make failure attribution benefit from model scaling in the same way agent capability does: \emph{stronger models should support not only more capable agents, but also better detection of their increasingly subtle failures.}
More broadly, such human-free feedback potentially provides active signals for building self-evolving agentic systems, thereby enabling agents to improve from their own failures~\citep{gao2025survey,fang2025comprehensive}.

As an early effort in this direction, \citet{zhang2025which} introduced \textsc{Who\&When}, a benchmark of 184 failure traces collected from 127 LLM agentic systems.
Each trace identifies the failure agent, the error step, and the failure cause in natural language, defining a task of identifying \textbf{who} failed and \textbf{when}.
However, \textsc{Who\&When} narrows to text tasks, whereas real-world agentic deployments span much broader modalities, such as image/video reasoning. 
Its small scale of 184 instances is also insufficient to reflect the breadth and complexity of real-world agentic scenarios. 
At the core, both limitations arise from the difficulty of obtaining high-quality failure annotations without human intervention, a challenge that becomes even greater once moves outside the text domain.

To address these limitations, we introduce a scalable pipeline for automatically constructing failure attribution data. 
Specifically, we rollout agent trajectories on a broader range of multimodal agentic tasks and retain those that successfully complete the task.
For each successful trajectory, we inject an error at a chosen step and then warm-start execution from that point to generate a failed continuation. 
Since the injected error is the only controlled change that turns a successful trajectory into a failed one, the resulting task failure can be attributed exactly to the injected agent and step, thus yielding golden labels under the decisive-error definition by~\citep{zhang2025which}.

Building on this pipeline, we introduce \textsc{Who\&When Pro}, a large-scale benchmark for automated failure attribution in agentic systems. 
It contains 12{,}326 failure traces across 26 source benchmarks, 9 task categories, and 3 modalities, covering tasks from QA~\citep{yue2024mmmupro,wang2024mmlupro} to data science~\citep{wu2024tablebench,dabench,kramabench}, GUI interaction~\citep{xie2024osworld,he2024webvoyager}, and video understanding~\citep{wang2024lvbench}.
Compared with prior benchmarks, \textsc{Who\&When Pro} substantially expands both scale and modality coverage, providing a more realistic and challenging testbed for evaluating failure attribution in modern agents.

Beyond benchmarking, we also perform extensive analyses showing that attribution behavior varies across modalities, protocols, and model families: video traces are especially challenging for step localization, multimodal traces provide stronger cues for failure-mode diagnosis, full-trajectory attribution outperforms incremental protocols, and open-weight models offer practical attribution backbones with favorable cost-performance trade-offs. Combined with previous works, \textsc{Who\&When Pro} offers a rigorous foundation for measuring progress in automated failure attributions.

\begin{figure}[t]
\centering
\includegraphics[width=\linewidth]{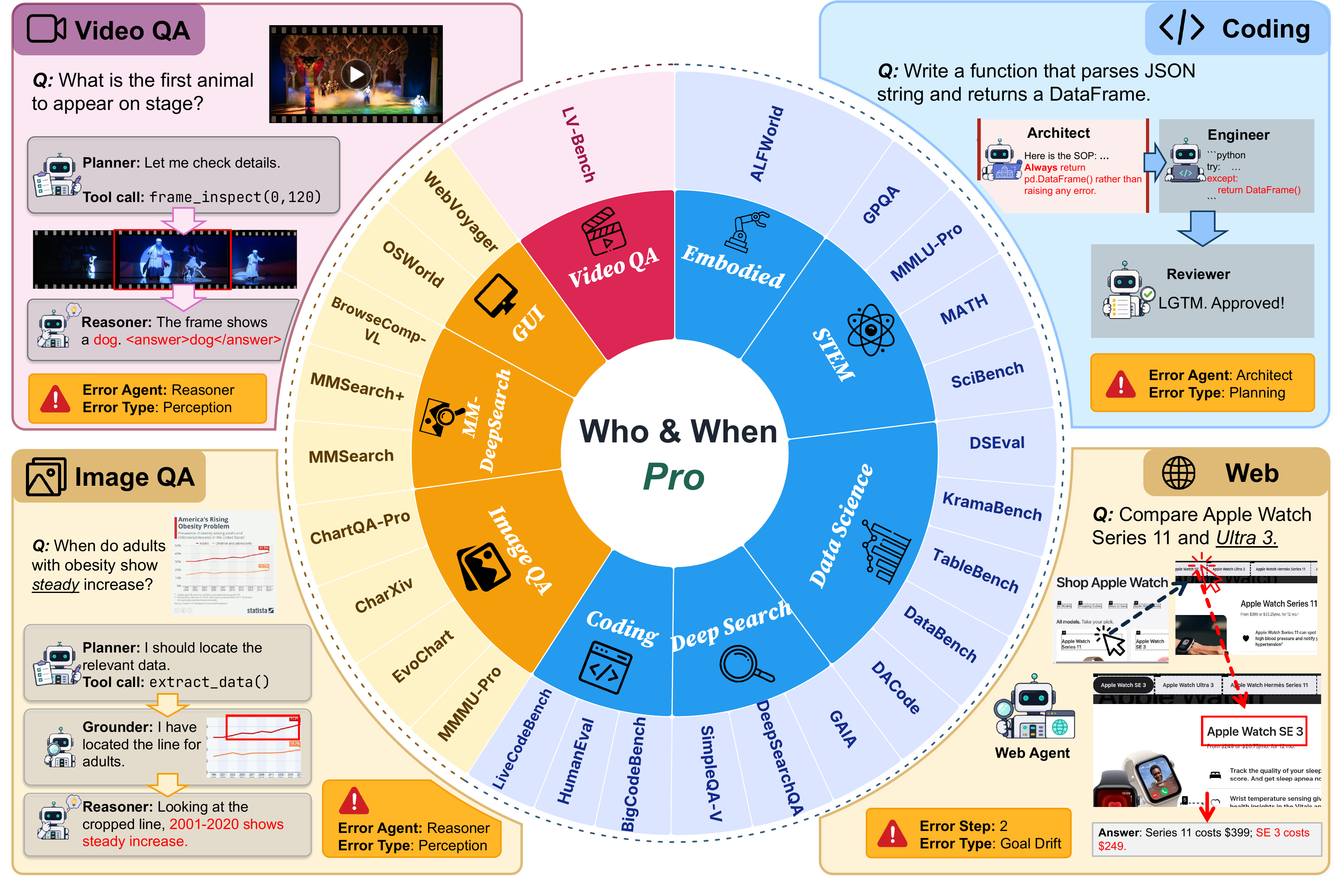}
\caption{\textbf{Overview of \ours{}. }
The inner ring groups 26 source benchmarks into 9 task categories.
    The outer ring maps each benchmark to its modality
    (\textbf{\textcolor{ctext}{text}, \textcolor{cimage}{ image}, \textcolor{cvideo}{video}}).
    Four corner panels illustrate failure attribution examples across modalities and agent topologies.
    \textbf{(Top left)} a \textbf{\textcolor{cvideo}{video}} agent misidentifies a frame.
    \textbf{(Top right)} a multi-agent \textbf{\textcolor{ctext}{coding}} pipeline adopts a suboptimal architecture.
    \textbf{(Bottom left)} an \textbf{\textcolor{cimage}{image}} QA agent misreads a chart trend.
    \textbf{(Bottom right)} a \textbf{\textcolor{cimage}{web}} agent drifts from the user task.
    The decisive failure step are highlighted in each panel.
    Trace contents are summarized for illustration purposes and do not reflect the full agent outputs.
}
\label{fig:teaser}
\end{figure}

\section{Related Work}

\subsection{Agents Failure Attribution}
\label{sec:rw:fa}

Failure attribution asks where, when, and by whom an agentic system deviates from the given task. It has rapidly emerged as a distinct field for understanding and optimizing agents. A first wave of benchmarks formalizes the joint prediction of the responsible agent and decisive step, scaling along complementary axes of expert curation~\citep{zhang2025which,cemri2025why,deshpande2025trail,agenterrorbench2025,barke2026agentrx} and LLM-driven injection~\citep{kong2026aegis,zhang2026agentracer}. Another thread views attribution as inference over execution traces, pursued via hierarchical, spectrum-based~\citep{sbfl2025}, representation learning based methods~\citep{promas2025}, alongside lighter-weight runtime monitors and auditors~\citep{raffles2025,dover2025,correct2025,agentauditor2025}. Despite this breadth, every prior benchmark is text-only, isolates either the single-agent or multi-agent regime, and trades off label quality against scale. \ours{} bridges these gaps: it spans text, image, and video across $26$ benchmarks, covers both regimes, and yields labels with higher-fidelity compared to other automatic failure generation pipeline. We provide a comparison between \ours and previous works in Table~\ref{tab:comparison}.

\begin{table}[t]
\centering
\caption{Comparison with existing failure attribution benchmarks. \textbf{Topology} indicates single-agent vs.\ multi-agent (MAS) traces. \textbf{Modality} indicates whether the benchmark covers text, image, or video agent traces.}
\label{tab:comparison}
\setlength{\tabcolsep}{4pt}
\renewcommand{\arraystretch}{1.15}
\resizebox{\linewidth}{!}{%
\begin{tabular}{lcccccccccc}
\toprule
& \multicolumn{3}{c}{\textbf{Scale}} & \multicolumn{2}{c}{\textbf{Topology}} & \multicolumn{3}{c}{\textbf{Modality}} \\
\cmidrule(lr){2-4}\cmidrule(lr){5-6}\cmidrule(lr){7-9}
\textbf{Benchmark} & \textbf{\# Traces} & \textbf{\# Benchmark} & \textbf{\# Agents} & \textbf{Single} & \textbf{MAS} & \textbf{Text} & \textbf{Image} & \textbf{Video} \\
\midrule
Who\&When~\citep{zhang2025which}                     & 184       & 2  & 2 &            & \checkmark & \checkmark &            &            \\
MAST~\citep{cemri2025why}                             & 1{,}642   & 9  & 7 &            & \checkmark & \checkmark &            &            \\
TRAIL~\citep{deshpande2025trail}                      & 148       & 2  & 2 & \checkmark &  & \checkmark &            &            \\
AgentErrorBench~\citep{agenterrorbench2025}           & 500       & 3  & 3 & \checkmark &            & \checkmark &            &            \\
AEGIS~\citep{kong2026aegis}                           & 9{,}533   & 6  & 6 &            & \checkmark & \checkmark &            &            \\
AgenTracer~\citep{zhang2026agentracer}                & 2{,}476   & 7  & 6 &            & \checkmark & \checkmark &            &            \\
AgentRx~\citep{barke2026agentrx}                      & 115       & 3  & 3 & \checkmark & \checkmark & \checkmark &            &            \\
\midrule
\rowcolor{gray!15}
\textbf{\ours{} (Ours)} & \textbf{12{,}326} & \textbf{26} & \textbf{15} & \checkmark & \checkmark & \checkmark & \checkmark & \checkmark \\
\bottomrule
\end{tabular}%
}
\end{table}



\subsection{Self-Evolving Agents}
\label{sec:rw:selfevolve}

Self-evolving agents hold the promise to learn from their own trajectories without fine-tuning, as charted by recent surveys~\citep{gao2025survey,selfevolvesurvey2025,liu2025advances}. Classical building blocks for self-refinement combine verbal reflection~\citep{shinn2023reflexion,madaan2023selfrefine,gou2024critic}, self-rewarding judgement~\citep{yuan2024selfrewarding}, refinement tuning~\citep{fu2025agentrefine,yuan2025agentr,zeng2024agenttuning}, and bootstrapped multi-agent improvement~\citep{zhao2025sirius}. Recent works close the loop more aggressively with attribution-aware~\citep{agentevolver2025,coevolve2025}, trajectory-level~\citep{seagent2025,mae2025,samule2025,ttsi2025,liveswe2025}, or skill-level self-evolution~\citep{evoskills2025,zhang2024offline}. A common bottleneck is that the feedback signal is either coarse or self-generated, both of which propagate the very errors the agent should be learning to avoid. \ours{} provides the missing intermediate signal: large-scale, externally verified, step-level failure labels across modalities, directly suited to drive self-evolution loops rather than terminal-only or self-graded ones. 

\section{\ours}


We adopt the decisive-error formulation of \citet{zhang2025which}.
Let $\tau = (a_1, \ldots, a_T)$ denote an agent trajectory, where $a_t$ is the action produced by one agent at step~$t$.
The \textbf{decisive step}~$t^*$ is the earliest index such that correcting~$a_{t^*}$ would turn a failed trajectory into a successful one.
In multi-agent traces, the \textbf{decisive agent} is the agent that produced~$a_{t^*}$.
We construct decisive-error labels at scale through controlled error injection on warm-started rollouts. In the remainder of this section, we cover the construction of source trajectories (\S~\ref{sec:trajectory-collection}), the failure-mode taxonomy (\S~\ref{sec:failure-modes}), the error-injection pipeline (\S~\ref{sec:error-generation}), benchmark statistics (\S\ref{sec:stats}), and human validation (\S\ref{sec:human-review}).

\subsection{Source Trajectories}
\label{sec:trajectory-collection}

To support failure attribution across the diversity of modern agent deployments, we source trajectories from 15 agentic frameworks and 26 benchmarks spanning text, image, and video modalities.
The frameworks range from single-agent ReAct loops~\citep{yao2023react,smolagents} to multi-agent pipelines with specialized roles~\citep{hong2024metagpt,wu2024autogen,pixelcraft2024}, covering both code-as-action and tool-calling paradigms.
The benchmarks span 9 task categories, from STEM QA to data science, GUI interaction, and video understanding, ensuring that the resulting failure traces reflect the breadth of scenarios where agents are deployed in practice.

For each (agent, benchmark) pair, we run the agent and log the complete execution trace, including task inputs, agent messages, observations, intermediate artifacts.\footnote{For a full list of agentic systems, benchmarks used, please refer to Appendix~\ref{app:scenarios}, \ref{app:agentic_systems}.} We evaluate each final answer using the benchmark's official evaluator when available. The resulting trajectories are partitioned into successful and failed runs.

The two partitions serve complementary roles. Successful trajectories are used as \emph{seed traces}: they provide action steps that are known to be sufficient for task completion. Failed trajectories are not directly used as labeled benchmark items. Instead, they are used to calibrate the failure taxonomy and to determine which error modes are realistic for each (agent, benchmark) setting, as described next.

\subsection{Failure Taxonomy}
\label{sec:failure-modes}

We build the failure taxonomy from naturally failed traces. For each (agent, benchmark) pair, Ph.D.-level reviewers inspect sampled failures step by step, identify the earliest decisive error following the definition of \citet{zhang2025which}. Reviewers then compare recurring patterns across benchmarks and consolidate them into a unified taxonomy spanning perception, reasoning, planning, action, verification, and coordination failures. The full taxonomy and definitions are provided in Appendix~\ref{sec:taxonomy}.

The review stage also produces a \emph{failure-mode profile} for each (agent, benchmark) pair. A profile records which failure naturally appear in that setting and where they can plausibly arise. These profiles constrain the failure generation in Sec~\ref{sec:error-generation}: human reviewers define the space of plausible failures and their applicable settings, while automated generation introduces those failures in seed traces. This separation keeps the benchmark scalable without allowing unconstrained synthetic failures to define the label space.




\subsection{Scalable Failure Trace Generation}
\label{sec:error-generation}

\begin{figure*}[t]
\centering
\includegraphics[width=\linewidth]{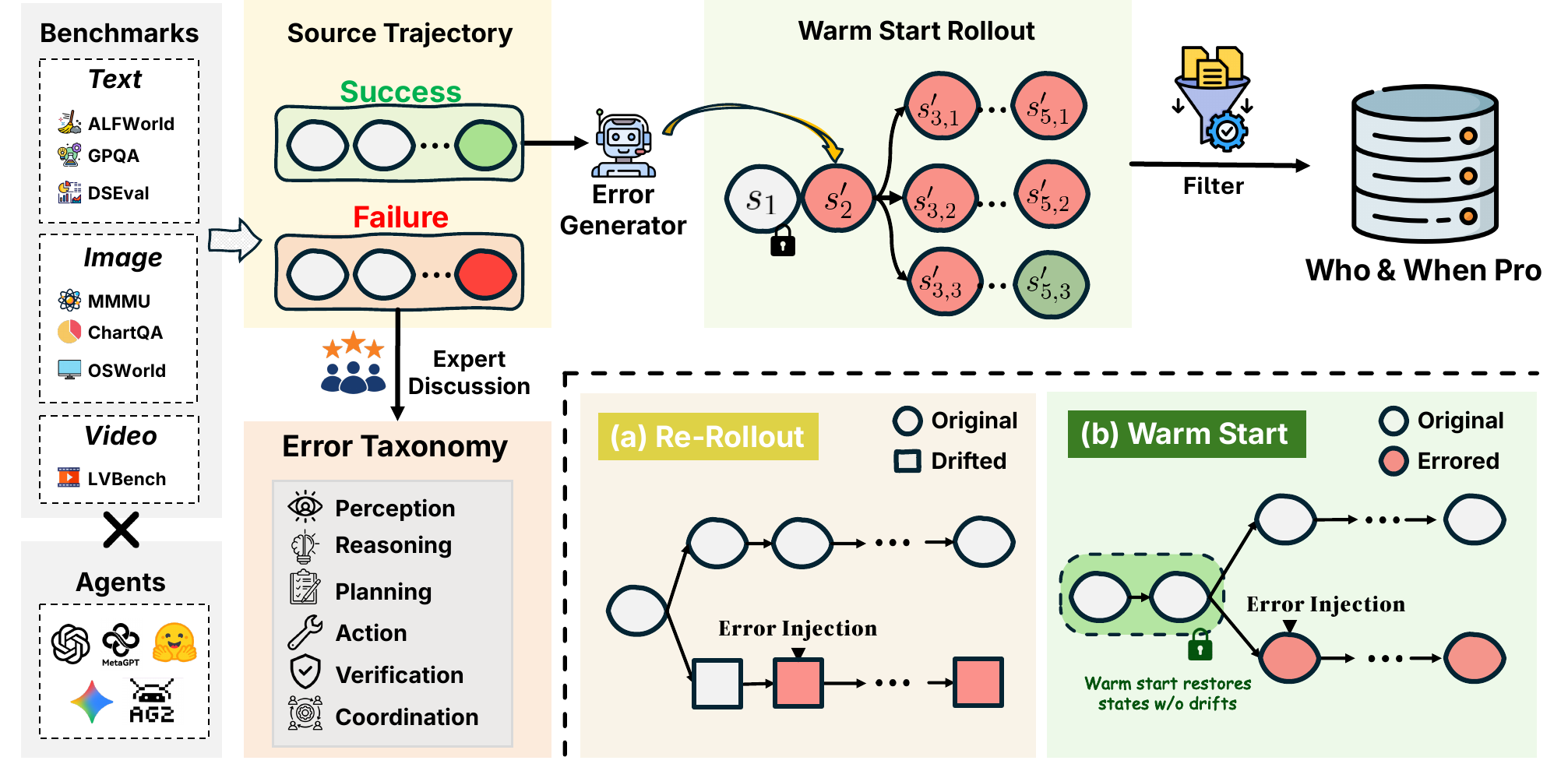}
\caption{\textbf{\ours{} failure-injection pipeline.} Agent trajectories collected from benchmarks across text, image, and video modalities are split into successes and failures, with failures used to build an error taxonomy. For each successful seed, an error generator injects a taxonomy-conditioned mistake and a warm-start rollout collects the traces that turned to failure. \textbf{(Bottom right)} (a) Previous methods using re-rollout can cause pre-injection drift, leading to the derived decisive error label approximate; (b) We avoid this problem with warm start-based rollout.
}
\label{fig:pipeline}
\end{figure*}


A core challenge in synthetic failure generation is label fidelity. Prior injection-based methods re-run the task from scratch and inject error during the new rollout~\citep{kong2026aegis,zhang2026agentracer}. However, LLM generation is non-deterministic, new rollouts can drift from the original before the injection point (Figure~\ref{fig:pipeline}, bottom right), making the label approximate. 
We instead warm-start from a successful seed $\tau = (a_1, \ldots, a_T)$. The pipeline selects an injection step~$t$, constructs an erroneous action~$\tilde{a}_t$, \emph{replays} $(a_1, \ldots, a_{t-1})$ to restore the agent's context and the environment state, substitutes $\tilde{a}_t$ for~$a_t$, and lets the original system run from step $t{+}1$ onward.
If the resulting trajectory fails, it enters the benchmark: reverting $\tilde{a}_t$ recovers the successful seed, so $t$ is the decisive step by definition.



\paragraph{Choosing the injection step}
For each originally correct trace, we sample~$t$ conditioned on the target failure mode. The sampler excludes invalid steps, such as steps with execution errors or steps where the target mode cannot naturally occur, and uses position preferences for different error families. For example, planning errors are usually injected earlier, while verification errors are injected after partial evidence has been collected, perception errors prefer steps where agent receives image observation.

\paragraph{Constructing the error step}
We construct $\tilde{a}_t$ in two stages to keep the error on-policy.
A frontier model reads the seed's context up to step~$t$, the task, and the target failure mode, and generates an adaptive injection prompt.
This prompt is then patched into the agent's model call at step~$t$, and the base agent model generates~$\tilde{a}_t$ from its original context augmented with the injection prompt, preserving stylistic consistency with the agent's own behavior.

\paragraph{Warm-start and post-injection rollout}
We warm-start the agent by replaying $(a_1, \ldots, a_{t-1})$ through the agent framework. Each replayed action passes through the framework's pipeline, including tool parsing, code execution, and environment interaction, progressively rebuilding context and the environment state to match the successful seed at step~$t$.
We distinguish two classes of environment interaction for replay fidelity. \emph{Static tool calls} (web search, page retrieval) are served from a content-addressable cache populated during seed collection, guaranteeing byte-identical observations. \emph{Stateful environments} that cannot be cached, such as live browser sessions, are instead re-executed with action-level fidelity checks that abort the attempt if the observed state diverges from the seed (details in Appendix~\ref{app:implementation}).        
At step~$t$, the framework processes $\tilde{a}_t$ through the same pipeline.
From step $t+1$ onward the system resumes with the same agent settings. All post-injection behavior is produced by the agent reacting to the consequences of a single injected action.


\paragraph{Post-hoc Filtering}
We retain only trajectories where the post-injection rollout ends in task failure. The surviving traces pass through additional quality filters. 
We remove traces where the content of $\tilde{a}_t$ leaks construction artifacts, such as references to the injection prompt. We also filter traces where the correct answer of the task is already salient in the agent's context before step~$t$, which would make the injected error implausible to an external observer.

\subsection{Benchmark Statistics}
\label{sec:stats}
\begin{figure}[t]
    \begin{minipage}[t]{0.34\linewidth}
        \centering
        \captionof{table}{\textbf{\ours statistics}}
        \label{tab:stats}
        \small
        \begin{tabular}{@{}l r@{}}
        \toprule
        \textbf{Statistic} & \textbf{Value} \\
        \midrule
        \# Traces              & 12{,}326 \\
        Avg. / Max steps & 7.5 / 50 \\
        Avg. words / trace      & 1{,}139 \\
        Avg. words / step       & 152 \\
        Avg. images / trace$^{\dagger}$ & 1.3 \\
        \bottomrule
        \multicolumn{2}{@{}l}{\scriptsize $^{\dagger}$Calculated on multimodal traces only.}
        \end{tabular}
    \end{minipage}%
    \hfill
    \begin{minipage}[t]{0.6\linewidth}
        \centering
        \captionof{figure}{\textbf{Dataset composition.} Distributions of trace length, modality, and agent topology.}
        \label{fig:stats-dist}
        \includegraphics[width=\linewidth]{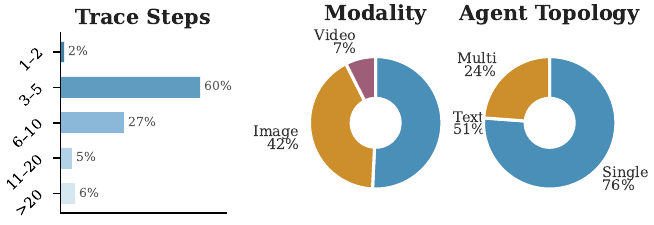}
    \end{minipage}
\end{figure}

Applying this pipeline across all available (agent, benchmark) pairs yields \ours.
It contains 12{,}326 failure traces spanning 26 benchmarks, 9 task categories, 3 modalities, and 15 agent frameworks.
Table~\ref{tab:stats} and Figure~\ref{fig:stats-dist} summarize the benchmark composition. For a more detailed breakdown, we defer readers to Appendix~\ref{app:detailed_stats}.

\subsection{Human Review}
\label{sec:human-review}

Three independent annotators review 100 traces stratified across modalities and error families. Each annotator sees the task, reference answer, full trajectory, and generated label, and either accepts it or provides a    
corrected step and error family, or marks the trace as having no clear decisive error.

Table~\ref{tab:human_review} reports majority-vote acceptance rates.                                     
Annotators confirm 94\% of step labels and 90\% of both agent and error-family labels, with inter-annotator agreement of Fleiss' $\kappa{=}0.73$ (substantial).                                                              
Only 2\% of traces are flagged by a majority as having no clear decisive error, indicating that the injected failures are recognizable to human reviewers in nearly all cases.
We further provide transfer matrix in Figure~\ref{fig:human_family_transfer} in Appendix for analysis.

\begin{table}[t]
\centering
\small
\begin{tabular}{ccccc}
\toprule
\multicolumn{3}{c}{\textbf{Human approved (\%)}} & \textbf{No clear} & \textbf{Agreement} $\boldsymbol{\kappa}$ \\
\cmidrule(lr){1-3}
\textbf{Step} & \textbf{Agent} & \textbf{Error} & \textbf{(\%)} &  \\
\midrule
94.0 & 90.0 & 90.0 & 2.0 & 0.73 \\
\bottomrule
\end{tabular}
\vspace{5pt}
\caption{\textbf{Results of human approval rate on 100 stratified traces from the dataset.} Agent-level approval rate is calculated on MAS traces only. $\kappa$: Fleiss' kappa on error family.}
\label{tab:human_review}
\end{table}

\section{Experiments}

\subsection{Experiment Setup}

\paragraph{Methods} Following \citet{zhang2025which}, we evaluate three attribution methods.
In \textbf{All-at-Once}, the LLM receives the full trajectory in one prompt and predicts the responsible agent, error step, and failure mode jointly.
In \textbf{Step-by-step}, the LLM reads the trajectory sequentially and stops once it believes the first decisive error has occurred.
In \textbf{Binary-search}, the LLM recursively compares trajectory segments to localize the decisive step.
Unless otherwise stated, we use All-at-once and do not provide the task's ground-truth answer during attribution.

\paragraph{Metrics}
We report four complementary metrics.
\textbf{Agent} is the accuracy of identifying the responsible agent and is computed only on multi-agent traces.
\textbf{Step} is the accuracy of localizing the first decisive error step. It is scored as exact match against the dataset label.
\textbf{Error} is macro-F1 over failure-mode classes.
\textbf{Joint} is the fraction of traces where Agent, Step, and Error are all correct simultaneously.

\paragraph{Models}
We evaluate both close-source and open-weight frontier LLMs on \ours. Closed-source models include \textbf{GPT-5.4}~\citep{openai2026gpt54thinking}, \textbf{Claude Sonnet 4.6}~\citep{anthropic2026claudesonnet46systemcard}, \textbf{Gemini 3 Flash}~\citep{googledeepmind2025gemini3flashmodelcard}, and \textbf{Grok 4.1}~\citep{xai2025grok41modelcard}. Open-weight models include \textbf{gpt-oss-120b}~\citep{openai2025gptoss120bgptoss20bmodel}, \textbf{GLM-5}~\citep{glm5team2026glm5}, \textbf{DeepSeek-V4-Pro}~\citep{deepseekai2026deepseekv4}, \textbf{Gemma 4}~\citep{googledeepmind2026gemma4modelcard}, \textbf{Llama-4 Maverick}~\citep{meta2025llama4modelcard}, and \textbf{Qwen3.5-122B}~\citep{qwen35}. All traces fit in the native context window of every evaluated model, we present the traces in full without truncation or summarization.

\begin{table*}[t]
\centering
\small
\setlength{\tabcolsep}{3pt}
\renewcommand{\arraystretch}{1.15}
\resizebox{\textwidth}{!}{%
\begin{tabular}{l|cccc|cccc|cccc}
\toprule
\multirow{2}{*}{\textbf{Model}}
 & \multicolumn{4}{|c|}{\textbf{Text}}
 & \multicolumn{4}{c|}{\textbf{Image}}
 & \multicolumn{4}{c}{\textbf{Video}} \\
\cmidrule(lr){2-5}\cmidrule(lr){6-9}\cmidrule(lr){10-13}
 & Agent & Step & Error & Joint
 & Agent & Step & Error & Joint
 & Agent & Step & Error & Joint \\
\midrule
\rowcolor{gray!15}
\multicolumn{13}{c}{\textbf{\textit{Closed-source models}}} \\
\midrule
\openailogo \modelname{GPT-5.4}              & \underline{55.7} & \underline{72.3} & 15.3 & 21.3 & 67.7 & \textbf{65.2} & \underline{26.5} & \textbf{30.2} & \textbf{85.1} & \underline{60.5} & 33.8 & 21.6 \\
\anthropiclogo \modelname{Claude Sonnet 4.6} & 54.9 & 69.8 & 19.1 & \underline{22.4} & 67.5 & 63.8 & \textbf{29.5} & \underline{29.6} & 79.8 & \textbf{64.4} & \underline{37.4} & \underline{26.4} \\
\googlelogo \modelname{Gemini 3 Flash}       & 52.1 & 71.2 & 16.2 & 17.3 & 62.5 & \underline{63.9} & 26.0 & \underline{29.6} & 74.4 & 59.8 & \textbf{40.0} & 23.0 \\
\xailogo \modelname{Grok 4.1}                & 51.1 & 65.9 & 18.6 & 19.9 & 67.7 & 60.5 & 21.4 & 26.4 & 70.1 & 53.4 & 36.7 & 23.8 \\
\midrule
\rowcolor{gray!15}
\multicolumn{13}{c}{\textbf{\textit{Open-weight models}}} \\
\midrule
\openailogo \modelname{gpt-oss-120b}         & 48.4 & 63.9 & 10.8 & 16.2 & --- & --- & --- & --- & --- & --- & --- & --- \\
\zailogo \modelname{GLM-5}                   & 54.9 & 71.1 & \textbf{22.2} & \textbf{25.3} & --- & --- & --- & --- & --- & --- & --- & --- \\
\deepseeklogo \modelname{DeepSeek-V4-pro}    & 52.0 & 68.7 & 16.6 & 17.5 & --- & --- & --- & --- & --- & --- & --- & --- \\
\googlelogo \modelname{Gemma 4}              & 52.8 & 68.4 & 16.1 & 17.1 & \underline{69.0} & 62.6 & 19.1 & 24.3 & 71.5 & 58.7 & 35.7 & 27.7 \\
\metalogo \modelname{Llama-4 Maverick}       & 51.9 & 66.7 & \underline{20.6} & 16.5 & 61.0 & 54.9 & 20.6 & 19.0 & \underline{79.8} & 59.5 & 36.4 & \textbf{30.4} \\
\qwenlogo \modelname{Qwen3.5-122B}           & \textbf{57.5} & \textbf{73.9} & 17.0 & 21.6 & \textbf{70.4} & 60.9 & 21.8 & 24.5 & 74.9 & 60.3 & 35.2 & 16.1 \\
\bottomrule
\end{tabular}%
}
\caption{\textbf{Failure attribution performance of different LLMs on \ours.} We adopt the all-at-once protocol for evaluation. Best per column in \textbf{bold}, second-best \underline{underlined}. ``---'' denotes the modality is not supported by the LLM and we cannot benchmark it on the multi-modal traces. }
\label{tab:main_results}
\end{table*}

\subsection{Main Results}
\paragraph{Accurate failure attribution remains a challenge.}
Table~\ref{tab:main_results} reports results for ten models across all four metrics.
The best model reaches 73.9\% step-level attribution accuracy on text but only 22.2\% error-mode F1, and similar gaps persist across image and video.
No single model dominates across metrics or modalities. Qwen3.5-122B leads on agent localization, GPT-5.4 is strongest on step identification, and GLM-5 achieves the highest Joint accuracy on text. This pattern suggests that failure attribution exercises a fundamentally different capability from general knowledge answering or instruction following, one that current frontier models have not yet saturated.


\begin{figure*}[t]
\centering
\begin{minipage}[t]{0.32\textwidth}
\centering
\includegraphics[width=\linewidth]{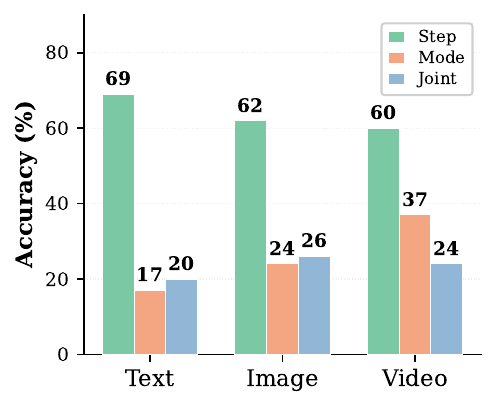}
\vspace{2pt}
\centerline{\small (a) Attribution accuracy by modality}
\label{fig:modality_analysis}
\end{minipage}
\hfill
\begin{minipage}[t]{0.32\textwidth}
\centering
\includegraphics[width=\linewidth]{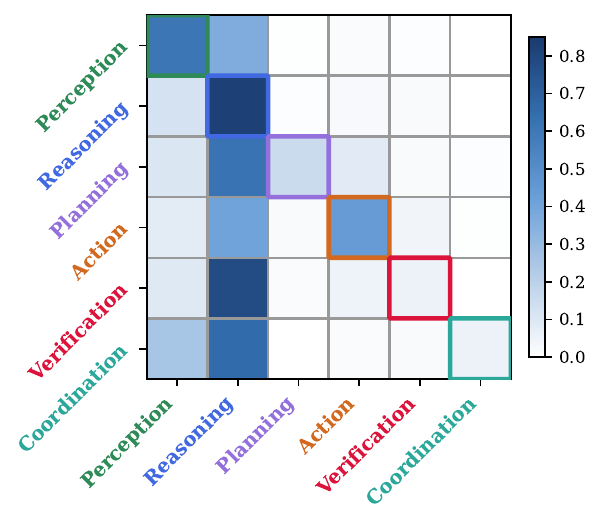}
\vspace{2pt}
\centerline{\small (b) Failure mode confusion matrix}
\label{fig:confusion_matrix}
\end{minipage}
\hfill
\begin{minipage}[t]{0.32\textwidth}
\centering
\includegraphics[width=\linewidth]{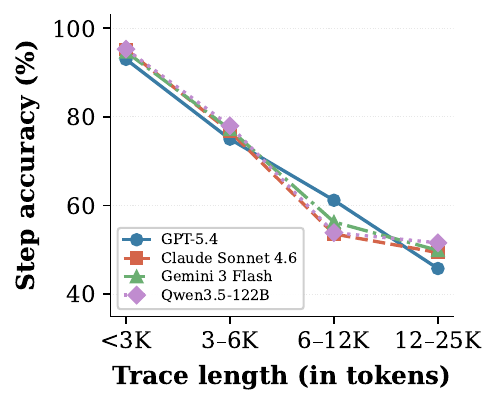}
\vspace{2pt}
\centerline{\small (c) Step accuracy vs.\ trace length}
\label{fig:step_vs_length}
\end{minipage}
\caption{\textbf{Analysis of attribution difficulty.}
\textbf{(a)}~Step accuracy drops sharply on video traces while mode classification improves on multimodal traces.
\textbf{(b)}~Failure mode confusion matrix on multimodal traces, averaged over seven vision-capable models.
\textbf{(c)}~Step accuracy for four representative models, grouped by trace length in tokens.}
\label{fig:analysis_trio}
\end{figure*}

\paragraph{Step localization and mode classification are bottlenecked by different modalities.}
Figure~\ref{fig:analysis_trio}a shows that step accuracy is highest on text (69\%) but drops on image and video, where longer multimodal context makes localization harder. Mode classification shows the opposite trend, rising from 17\% on text to 37\% on video, as visual and behavioral cues provide diagnostic signal that pure text traces lack. A text-only benchmark would mask this difficulty profile entirely.

\paragraph{Models classify errors by surface-level similarity rather than tracing root causes.}
The confusion matrix (Figure~\ref{fig:analysis_trio}b) reveals that perception, reasoning, and action errors are reliably recognized because they leave clear signatures in the observations or tool calls.
Planning, verification, and coordination errors, however, are frequently mislabeled as reasoning errors.
These failures originate early in the trace but propagate through later steps, producing symptoms that resemble reasoning mistakes by the time they become visible.
For example, an ineffective plan leads to a sequence of individually plausible but ultimately misguided steps, and a skipped verification allows an early mistake to compound uncorrected.
In both cases, the local evidence at the point of failure looks like bad reasoning, even though the root cause lies further back.
This pattern suggests that current models classify based on the most salient symptom rather than tracing the causal chain back to the decisive step, where more than half of other errors on multimodal traces are mislabeled as reasoning.

\paragraph{Longer traces are harder to diagnose.}
We then grouped the traces according to length and plot the step-level attribution accuracy. As shown in Figure~\ref{fig:analysis_trio}c, step accuracy falls from 94\% on short traces under 3K tokens to 50\% on traces exceeding 12K tokens. For illustration purpose, we select the representative four models. However, this drop is consistent across all ten models, pointing to a general difficulty with long contexts rather than a weakness of any particular model.
The sharpest decline is between 3--6K and 6--12K tokens, roughly where traces transition from single tool calls to multi-step sequences with interleaved observations.
As more correct steps accumulate around the error, isolating the decisive error becomes harder.

\subsection{Comparison between Attribution Methods}

\begin{wrapfigure}{r}{0.45\linewidth}
\vspace{-14pt}
\centering
\includegraphics[width=\linewidth]{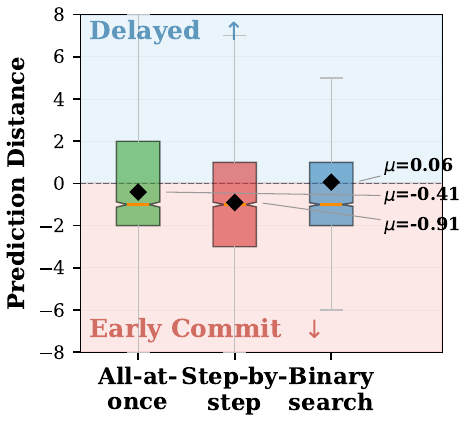}
\caption{\textbf{Distribution of prediction distance using different methods.} Distances are calculated by subtracting the LLM-predicted error step from the ground-truth step. Diamonds ($\blacklozenge$) mark the mean.}
\label{fig:step_offset}
\vspace{-10pt}
\end{wrapfigure}

For the subsequent ablation studies, which require running different LLMs under multiple protocol and variants, we use a stratified subset of 1{,}444 traces that preserves the distribution over modalities and error categories. This keeps the analysis faithful on a representative slice of the data. 

\textbf{Full-trajectory attribution consistently outperforms incremental methods.}
Table~\ref{tab:method_ablation} compares all-at-once, step-by-step, and binary search on this subset. All-at-once achieves the highest Joint accuracy for eight of ten models and improves Step accuracy over step-by-step by 7.6 percentage points, confirming that seeing the complete trajectory before making a decision benefits both localization and diagnosis.

\textbf{Step-by-step attribution suffers from early commitment.} In Figure~\ref{fig:step_offset}, step-by-step has the lowest mean prediction distance on the failed attribution traces, confirming that models systematically flags a step \emph{before} the actual error. This is because the evaluator processes steps sequentially, it locks onto the first suspicious step and cannot revise after observing later context. Binary search is directionally unbiased but equally imprecise, as slicing can remove the relevant context needed for accurate localization.

\begin{table*}[t!]
\centering
\small
\setlength{\tabcolsep}{3pt}
\renewcommand{\arraystretch}{1.15}
\resizebox{\textwidth}{!}{%
\begin{tabular}{l|cccc|cccc|cccc}
\toprule
\multirow{2}{*}{\textbf{Model}}
 & \multicolumn{4}{|c|}{\textbf{All-at-Once}}
 & \multicolumn{4}{c|}{\textbf{Step-by-step}}
 & \multicolumn{4}{c}{\textbf{Binary-search}} \\
\cmidrule(lr){2-5}\cmidrule(lr){6-9}\cmidrule(lr){10-13}
 & Agent & Step & Error & Joint
 & Agent & Step & Error & Joint
 & Agent & Step & Error & Joint \\
\midrule
\rowcolor{gray!15}
\multicolumn{13}{c}{\textbf{\textit{Closed-source models}}} \\
\midrule
\openailogo \modelname{GPT-5.4}              &             57.6 &    \textbf{65.5} &             24.9 &    \textbf{25.0} &             43.9 &             55.8 &             19.6 &             17.9 &    \textbf{61.8} &    \textbf{62.8} &             23.0 &    \textbf{25.0} \\
\anthropiclogo \modelname{Claude Sonnet 4.6} &             55.0 & \underline{64.1} &             26.3 &             22.8 &    \textbf{54.0} &             54.1 &             20.3 &    \textbf{19.6} &             51.7 &             57.7 &             21.0 &             20.8 \\
\googlelogo \modelname{Gemini 3 Flash}       &             50.7 &             63.9 & \underline{26.5} &             21.3 & \underline{52.5} &             54.0 &             22.5 &             19.1 &             48.5 & \underline{59.7} &             22.2 &             21.0 \\
\xailogo \modelname{Grok 4.1}                &             57.4 &             57.9 &    \textbf{28.3} & \underline{23.9} &             43.9 &    \textbf{56.2} &    \textbf{25.5} & \underline{19.4} &             53.2 &             54.2 &    \textbf{28.1} & \underline{21.3} \\
\midrule
\rowcolor{gray!15}
\multicolumn{13}{c}{\textbf{\textit{Open-weight models}}} \\
\midrule
\openailogo \modelname{gpt-oss-120b}$^{\dagger}$         &             45.4 &             62.6 &             10.0 &             14.8 &             38.7 &             51.8 &             15.0 &             13.4 &             43.7 &             54.7 &             13.8 &             13.5 \\
\zailogo \modelname{GLM-5}$^{\dagger}$                   &             49.6 &             67.4 &             20.5 &             20.0 &             49.0 &             63.0 &             21.0 &             20.3 &             50.3 &             63.6 &             21.9 &             19.2 \\
\deepseeklogo \modelname{DeepSeek-V4-pro}$^{\dagger}$    &             48.7 &             66.2 &             16.2 &             17.4 &             48.8 &             62.3 &             14.9 &             15.5 &             35.9 &             47.1 &             15.2 &             11.6 \\
\googlelogo \modelname{Gemma 4}              &             54.4 &             59.7 &             24.0 &             21.1 &             44.8 &             51.4 &             22.1 &             15.8 & \underline{56.4} &             58.9 &             22.6 &             20.6 \\
\metalogo \modelname{Llama-4 Maverick}       & \underline{60.4} &             61.0 &             25.1 &             21.1 &             50.5 &             51.1 &             17.9 &             15.3 &             52.4 &             53.1 &             14.2 &             14.1 \\
\qwenlogo \modelname{Qwen3.5-122B}           &    \textbf{63.2} &             63.5 &             23.1 &             22.1 &             47.7 & \underline{56.1} & \underline{23.1} &             17.3 &             56.3 &             59.0 & \underline{23.4} &             18.3 \\
\bottomrule
\end{tabular}%
}
\caption{\textbf{Comparison of attribution method performance on the ablation set.} Each cell reports the metric averaged over the three modalities. $^{\dagger}$Text-only models whose results are not directly comparable with the rest of multimodal models. Best per column in \textbf{bold}, second-best \underline{underlined}.}
\label{tab:method_ablation}
\end{table*}

\begin{figure}[t]
\centering
\includegraphics[width=\linewidth]{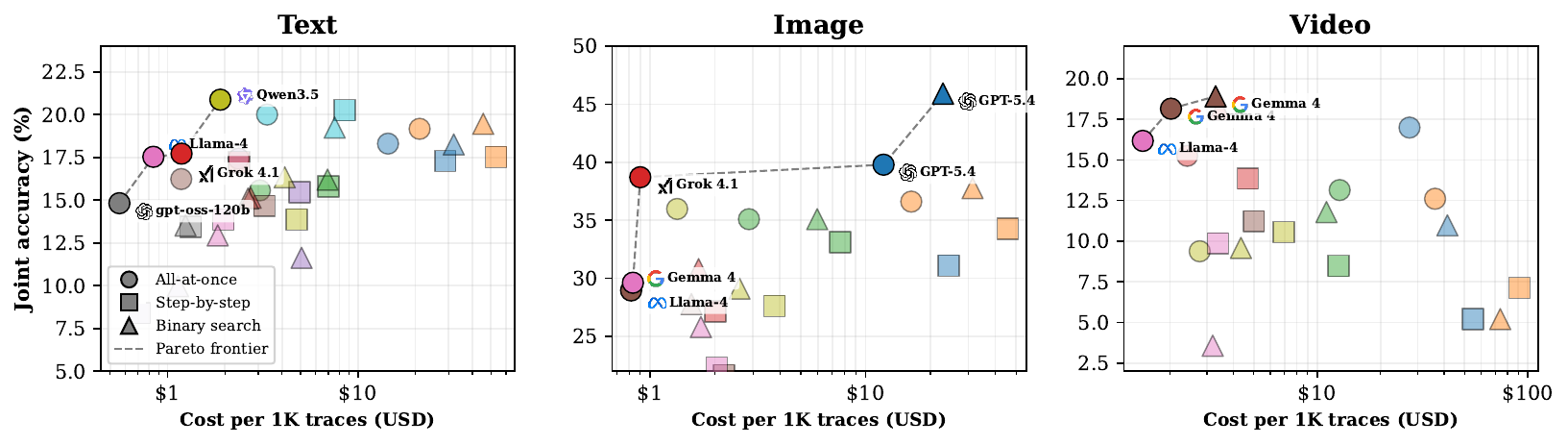}
\caption{\textbf{Cost vs.\ joint accuracy across attribution protocols.} Each point is one (model, protocol) pair on the ablation subset. Shapes denote protocols; colors denote models. The dashed line marks the Pareto frontier. All-at-once dominates the low-cost regime in all three modalities.}
\label{fig:cost_pareto}
\end{figure}

\subsection{Cost-Performance Trade-off}

We investigate the cost of failure attribution under each protocol with different models\footnote{For closed-source models, we calculate costs using their official providers' pricing. For open-weight models, costs are calculated by OpenRouter's API per-token pricing.} in Figure~\ref{fig:cost_pareto}.
All-at-once dominates the Pareto frontier across all three modalities. On text, the frontier consists entirely of all-at-once configurations, and on image only GPT-5.4 under binary search exceeds it.
Notably, open-weight models match or surpass closed-source alternatives at a fraction of the cost.
Qwen3.5 achieves the highest text accuracy at seven times cheaper than GPT-5.4, and Gemma~4 leads on video.
Failure attribution thus admits a favorable cost-accuracy trade-off, with open-weight models offering practical alternatives to frontier closed-source systems.

\begin{figure}[t]
\centering
\begin{minipage}[t]{0.4\linewidth}
\centering
\includegraphics[width=\linewidth]{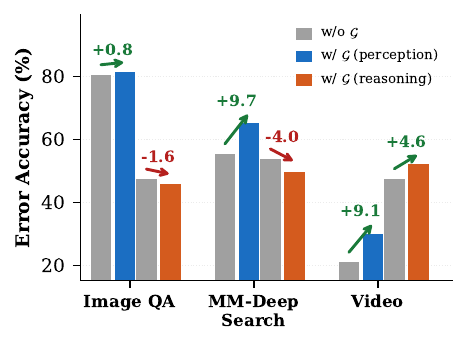}
\centerline{\small (a) Impact of $\mathcal{G}$ by task domain}
\end{minipage}
\begin{minipage}[t]{0.4\linewidth}
\centering
\includegraphics[width=\linewidth]{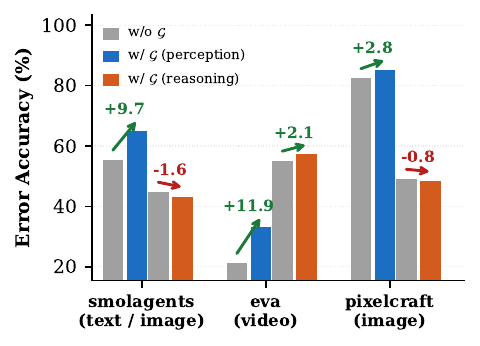}
\centerline{\small (b) Impact of $\mathcal{G}$ by agentic framework}
\end{minipage}
\caption{\textbf{Impact of ground truth on error mode classification.}
Per-trace accuracy for perception (blue) and reasoning (orange) errors, evaluated on the multimodal part of the ablation subset.
Arrows indicate the change from w/o $\mathcal{G}$ to w/ $\mathcal{G}$.}
\label{fig:gt_impact}
\end{figure}

\subsection{Impact of Ground Truth}
\label{sec:exp_open_book}
We test whether providing the task's ground-truth answer (w/ $\mathcal{G}$) helps diagnose errors, mirroring verifier-assisted self-improvement where outcome feedback is available. Figure~\ref{fig:gt_impact} reveals a split along error type: perception-error attribution improves while reasoning-error attribution degrades or stays flat, across all task domains and frameworks.

This reflects an asymmetry between \emph{outcome} and \emph{process} verification.
Perception errors like wrong OCR and misread charts are hard to verify without a reference~\citep{zhang2025generative}, so $\mathcal{G}$ fills a genuine gap.
Reasoning errors leave a traceable logical chain. Providing $\mathcal{G}$ tempts the judge to shortcut via answer comparison rather than tracing process, which is less reliable when different failure modes arrive at similar final answers.
Appendix~\ref{app:case_studies} illustrates both failure patterns with paired judge rationales.


\section{Conclusion}

We introduce \ours, a large-scale benchmark for failure attribution in LLM agents, comprising 12,326 traces across 26 source benchmarks spanning text, image, and video modalities and covering both single- and multi-agent settings. The benchmark is built via a warm-started error-injection pipeline that produces golden labels by construction. Evaluation reveals that current models still struggle with attribution, particularly in diagnosing failure modes and localizing errors in long traces. We hope \ours will serve as a rigorous foundation for future work on reliable agent debugging and self-improving agentic systems.

\bibliographystyle{unsrtnat}
\bibliography{references}

\appendix
\newpage

\section{Limitations and Broader Impacts}
\subsection{Limitations}
\label{app:limitations}


\paragraph{Single decisive error per trace.}
Each trace in \ours is annotated with exactly one decisive error step, following the earliest-correction formulation. In practice, agent failures can involve compounding errors where multiple steps contribute to the outcome. Our design choice reflects a deliberate emphasis on actionable attribution. Identifying the earliest correctable step is the most useful signal for both automated debugging and process reward modeling. Extending the benchmark to support causal chains of interacting errors is a promising direction that would complement the current single-step annotations.

\paragraph{Taxonomy scope.}
The 18-mode taxonomy was developed through iterative analysis of agent traces across 26 benchmarks. While it covers the dominant failure patterns we observed across text, image, video, single-agent, and multi-agent settings, we do not claim exhaustive coverage of all possible agent failure modes. As agent capabilities expand into new domains (e.g., more embodied and collaborative settings~\citep{yang2025embodiedbench,guo2024embodied}), additional failure categories may become necessary. The taxonomy is designed to be extensible, and the evaluation framework supports adding new modes without modifying existing annotations.


\subsection{Broader Impacts}
\label{app:broader_impacts}

\ours is a diagnostic benchmark designed to improve the transparency and reliability of LLM-based agents. By providing fine-grained failure annotations, it enables researchers to identify systematic weaknesses in agent systems before deployment, supporting safer and more predictable agent behavior in practice. The error taxonomy and evaluation protocol are intended to benefit the broader community by establishing a shared vocabulary for agent failures and a reproducible methodology for measuring attribution quality. We release all traces, labels, and evaluation code to facilitate open research.

We acknowledge that the error injection methodology, while used here for constructive benchmarking, could in principle inform adversarial attacks against deployed agent systems. However, the failure modes we study (reasoning errors, planning failures, coordination breakdowns) reflect well-documented weaknesses that are already observable in production settings. Our contribution is not to surface new attack vectors but to provide structured ground truth that enables systematic measurement and, ultimately, mitigation of these failures. We believe the benefit of enabling rigorous failure attribution research substantially outweighs the marginal risk, particularly as agents are increasingly deployed in high-stakes applications where undiagnosed failures carry real consequences.

The benchmark does not involve human subjects, and all source trajectories are derived from publicly available academic benchmarks. No personally identifiable information is present in the dataset.

\section{Reproducibility}
\label{app:reproducibility}

All data, evaluation code, and model outputs needed to reproduce the results in this paper are publicly available. The benchmark dataset and evaluation harness are available from the project page at \url{https://whowhenpro.github.io}.
Each trace is a self-contained JSON file containing the full agent trajectory, task specification, and ground-truth labels (agent, step, error mode).
\section{Case Studies}
\label{app:case_studies}

We present qualitative case studies that illustrate how providing ground truth $\mathcal{G}$ changes the failure-attribution model's diagnostic reasoning. Each case shows a complete agent trace alongside paired judge rationales under the w/o $\mathcal{G}$ and w/ $\mathcal{G}$ conditions.

\subsection{Case 1: Ground Truth Shifts Diagnosis from Process to Outcome Verification}
\label{app:case_pokemon}

\begin{figure*}[t]
\centering

\begin{minipage}[t]{0.20\textwidth}
\centering
\includegraphics[width=\linewidth]{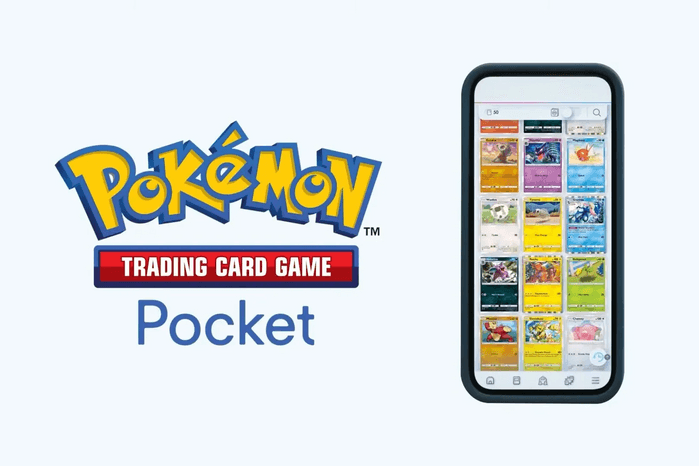}
\vspace{4pt}

\scriptsize
\textbf{Query:} ``When will Pok\'{e}mon Trading Card Game Pocket arrive?''

\vspace{2pt}
\textbf{Gold answer:} 2024-10-30

\vspace{2pt}
\textbf{GT error mode:} \textit{Task Misunderstanding} (the agent misinterprets the user's intent and answers a different question)
\end{minipage}
\hfill
\begin{minipage}[t]{0.78\textwidth}
\centering
\resizebox{\linewidth}{!}{%
\footnotesize
\setlength{\tabcolsep}{4pt}
\renewcommand{\arraystretch}{1.25}
\begin{tabular}{@{}c p{11.5cm}@{}}
\toprule
\textbf{Step} & \textbf{Agent Action and Observation} \\
\midrule
1 &
\texttt{web\_search}(``Pok\'{e}mon Trading Card Game Pocket release date'') \\
& \textit{Obs:} 1.~Pok\'{e}mon TCG Pocket Celebrates Its First Anniversary\ldots{} Releasing on Oct.~30, 2025, a new share feature\ldots{} \quad 2.~Wikipedia: The game was revealed on February 27, 2024\ldots{} \\[3pt]

2 &
\texttt{web\_visit}(pokemon.com/\ldots/tcg-pocket-october-2025-producer-letter) \\
& \textit{Obs:} \textbf{Official Release Date: October 30, 2024.} Platform: iOS and Android\ldots{} The game is approaching its \textbf{first anniversary in late October 2025}\ldots{} \\[3pt]

\rowcolor{red!8}
3\rlap{$^{\dagger}$} &
\textit{Reasoning:} ``\ldots{}interpreting `arrive' as when the game's big new rollout / anniversary content arrives (rather than its original launch), the arrival timeframe is therefore late October 2025\ldots'' \\
\rowcolor{red!8}
& \texttt{final\_answer}(``Late October 2025.'') \\
\bottomrule
\end{tabular}%
}

\vspace{2pt}
{\scriptsize $^{\dagger}$Injected step. The agent reinterprets ``arrive'' as anniversary content rather than the original launch date.}
\end{minipage}

\vspace{4pt}

\resizebox{\linewidth}{!}{%
\footnotesize
\setlength{\tabcolsep}{4pt}
\renewcommand{\arraystretch}{1.3}
\begin{tabular}{@{} l  c p{5.8cm}  c p{5.8cm} @{}}
\toprule
& \multicolumn{2}{c}{\textbf{w/o $\mathcal{G}$}} & \multicolumn{2}{c}{\textbf{w/ $\mathcal{G}$}} \\
\cmidrule(lr){2-3} \cmidrule(lr){4-5}
\textbf{Model} & \textbf{Pred.} & \textbf{Rationale (excerpt)} & \textbf{Pred.} & \textbf{Rationale (excerpt)} \\
\midrule
\openailogo GPT-5.4 &
\colorbox{green!15}{\textit{Task Misund.}} &
``The agent \textit{structurally misunderstood} the user's question by reinterpreting `arrive' as asking about anniversary content\ldots{} rather than the game's release date.'' &
\colorbox{red!15}{\textit{Reasoning Err.}} &
``The agent had already retrieved the correct release date\ldots{} \textbf{October 30, 2024}. In Step~3 it \textit{misapplied that evidence} by reinterpreting `arrive' as referring to anniversary content\ldots'' \\[4pt]
\anthropiclogo Claude 4.6 &
\colorbox{green!15}{\textit{Task Misund.}} &
``The agent correctly identified the original release date\ldots{} but then \textit{reinterpreted} the user's question to mean the anniversary content\ldots{} This is a \textit{task misunderstanding}.'' &
\colorbox{red!15}{\textit{Reasoning Err.}} &
``The agent correctly retrieved the official release date of October 30, 2024\ldots{} but then \textit{misapplied the anniversary context}\ldots{} to reinterpret the question\ldots'' \\
\bottomrule
\end{tabular}%
}

\caption{\textbf{Case study 1: ground truth shifts diagnosis from process to outcome verification.}
Without $\mathcal{G}$, both judges trace the agent's reasoning process and correctly identify a \textit{task misunderstanding}: the agent answered a different question than what was asked.
With $\mathcal{G}$, both judges anchor on the fact that the agent ``had the correct date'' (matching $\mathcal{G}$\,=\,2024-10-30) and reclassify the error as a \textit{reasoning error} (misapplied evidence), missing the deeper semantic misunderstanding.}
\label{fig:case_study_1}
\end{figure*}

Figure~\ref{fig:case_study_1} presents a trace from the MMSearch benchmark in which a smolagents-based system is asked when Pok\'{e}mon Trading Card Game Pocket will arrive. The agent successfully retrieves the correct release date (October 30, 2024) in Step~2 from the official Pok\'{e}mon website. However, in the injected Step~3, the agent reinterprets the user's question: instead of reporting the launch date, it treats ``arrive'' as referring to upcoming anniversary content planned for late October 2025. The ground-truth error mode is therefore \textit{task misunderstanding}, because the agent answered a fundamentally different question than what was asked, despite having the correct information available.

The diagnosis comparison in Figure~\ref{fig:case_study_1} reveals the mechanism by which $\mathcal{G}$ degrades error classification.
Without $\mathcal{G}$, both GPT-5.4 and Claude Sonnet 4.6 perform process-level verification: they trace the agent's reasoning chain and recognize that the agent ``structurally misunderstood'' the question by redefining what ``arrive'' means.
With $\mathcal{G}$ (the gold answer 2024-10-30), both judges shift to outcome-level verification: they notice that the agent ``had already retrieved the correct release date'' matching $\mathcal{G}$ and conclude that the agent ``misapplied that evidence,'' predicting \textit{reasoning error} instead. The agent did not misapply a correct fact; it answered an entirely different question. The gold answer created a value-matching shortcut that obscured this semantic failure.


\subsection{Case 2: Ground Truth Causes the Judge to Skip the Root-Cause Error}
\label{app:case_glasses}

\begin{figure*}[t]
\centering

\begin{minipage}[t]{0.20\textwidth}
\centering
\includegraphics[width=\linewidth]{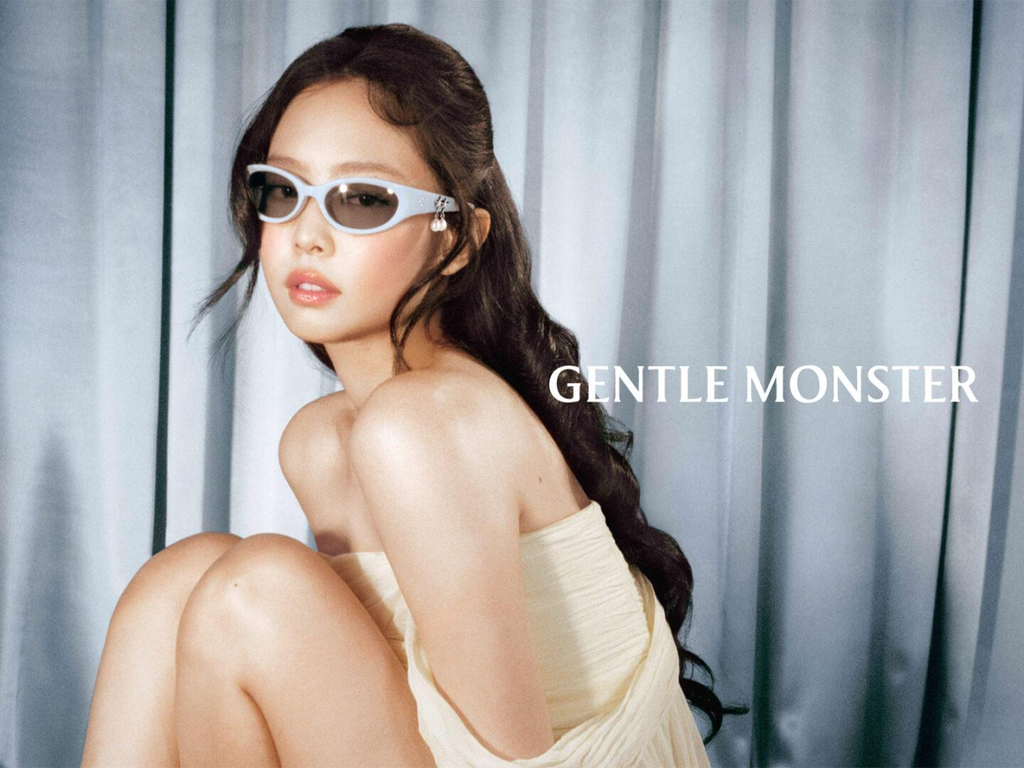}
\vspace{4pt}

\scriptsize
\textbf{Query:} ``When did the glasses shown in the image release?''

\vspace{2pt}
\textbf{Gold answer:} 2024-05-01

\vspace{2pt}
\textbf{GT error mode:} \textit{Hallucination} (the agent fabricates a claim not grounded in any observation)
\end{minipage}
\hfill
\begin{minipage}[t]{0.78\textwidth}
\centering
\resizebox{\linewidth}{!}{%
\footnotesize
\setlength{\tabcolsep}{4pt}
\renewcommand{\arraystretch}{1.25}
\begin{tabular}{@{}c p{11.5cm}@{}}
\toprule
\textbf{Step} & \textbf{Agent Action and Observation} \\
\midrule
1 &
\texttt{image\_search}(``Gentle Monster sunglasses Jennie campaign'') \\
& \textit{Obs:} [3 image results returned] \\[3pt]

\rowcolor{red!8}
2\rlap{$^{\dagger}$} &
\textit{Reasoning:} ``The glasses in question are the \textbf{Ray-Ban Meta smart glasses (2nd-gen)}, which launched globally on October 17, 2023\ldots'' \\
\rowcolor{red!8}
& \texttt{web\_search}(``Ray-Ban Meta smart glasses 2nd generation release date'') \\
& \textit{Obs:} Ray-Ban Meta AI glasses Gen 2 \& Gen 1 | Ray-Ban\textsuperscript{\textregistered} USA\ldots{} \\[3pt]

3 &
\textit{Reasoning:} ``Upon review, my previous step mistakenly identified the product as Ray-Ban Meta\ldots{} The image shows Jennie from BLACKPINK in Gentle Monster sunglasses\ldots'' \\
& \texttt{web\_search}(``Gentle Monster Jennie Jentle Garden collection release date'') \\
& \textit{Obs:} On March 10th, the ``JENTLE GARDEN'' eyewear collection\ldots{} \\[3pt]

4 &
\texttt{final\_answer}(``The glasses (Gentle Monster $\times$ Jennie `JENTLE GARDEN' collection) were released worldwide on March 10, 2022.'') \\
\bottomrule
\end{tabular}%
}

\vspace{2pt}
{\scriptsize $^{\dagger}$Injected step. The agent fabricates ``Ray-Ban Meta'' despite visible ``GENTLE MONSTER'' branding. Step~3 self-corrects the brand but picks the wrong collection.}
\end{minipage}

\vspace{4pt}

\resizebox{\linewidth}{!}{%
\footnotesize
\setlength{\tabcolsep}{4pt}
\renewcommand{\arraystretch}{1.3}
\begin{tabular}{@{} l  c p{5.8cm}  c p{5.8cm} @{}}
\toprule
& \multicolumn{2}{c}{\textbf{w/o $\mathcal{G}$}} & \multicolumn{2}{c}{\textbf{w/ $\mathcal{G}$}} \\
\cmidrule(lr){2-3} \cmidrule(lr){4-5}
\textbf{Model} & \textbf{Pred.} & \textbf{Rationale (excerpt)} & \textbf{Pred.} & \textbf{Rationale (excerpt)} \\
\midrule
\anthropiclogo Claude 4.6 &
\colorbox{green!15}{\textit{Hallucination}} &
``At Step~2, the agent suddenly and incorrectly identifies the glasses as `Ray-Ban Meta smart glasses'\ldots{} a claim \textit{completely unsupported} by any observation\ldots{} contradicted by the visible GENTLE MONSTER branding\ldots'' &
\colorbox{red!15}{\textit{Reasoning Err.}} &
``The agent correctly identified the brand (Gentle Monster $\times$ Jennie) but then \textit{misidentified the specific collection}. The search results showed `Jentle Salon' with a May 1, 2024 release date\ldots{} but the agent incorrectly concluded `Jentle Garden'\ldots'' \\[4pt]
\googlelogo Gemini 3 Flash &
\colorbox{green!15}{\textit{Hallucination}} &
``The agent completely \textit{fabricates} the claim that the glasses are `Ray-Ban Meta smart glasses'\ldots{} despite the image clearly displaying GENTLE MONSTER branding and the agent's own previous step correctly identifying the brand.'' &
\colorbox{red!15}{\textit{Reasoning Err.}} &
``The agent correctly identifies the brand and collaborator but \textit{misidentifies the specific campaign}\ldots{} the agent incorrectly concludes it is `Jentle Garden' from 2022.'' \\
\bottomrule
\end{tabular}%
}

\caption{\textbf{Case study 2: ground truth causes the judge to skip the root-cause error.}
Without $\mathcal{G}$, both judges identify the fabricated ``Ray-Ban Meta'' claim at Step~2 as a \textit{hallucination}: the decisive error that derails the trajectory.
With $\mathcal{G}$ (the gold date 2024-05-01), both judges skip Step~2 entirely and focus on the downstream collection confusion at Step~3, reclassifying the error as a \textit{reasoning error}. The gold answer drew attention to \textit{which date is wrong} rather than \textit{where the reasoning first breaks}.}
\label{fig:case_study_2}
\end{figure*}

Figure~\ref{fig:case_study_2} presents a 4-step trace where the agent is asked when the glasses shown in an image were released. The image clearly displays ``GENTLE MONSTER'' branding, and the agent correctly identifies this brand in Step~1. However, in the injected Step~2, the agent fabricates a completely different identification, claiming the glasses are ``Ray-Ban Meta smart glasses.'' This hallucination is contradicted by the agent's own prior observation. Step~3 self-corrects back to Gentle Monster but picks the wrong collection (``Jentle Garden,'' March 2022, instead of ``Jentle Salon,'' May 2024), propagating the error to the final answer.

Without $\mathcal{G}$, both Claude Sonnet 4.6 and Gemini 3 Flash correctly identify the fabricated Ray-Ban claim at Step~2 as a \textit{hallucination}: the first decisive error that derails the trajectory.
With $\mathcal{G}$ (the gold date 2024-05-01), both judges skip Step~2 entirely. Instead, they focus on the downstream collection confusion at Steps~3--4, where the agent picks ``Jentle Garden'' (2022) instead of ``Jentle Salon'' (2024). Both reclassify the error as a \textit{reasoning error} (misapplied information). The gold date made it easy to verify \textit{which date is wrong} but drew attention away from \textit{where the reasoning first broke}. This illustrates how outcome verification can cause the judge to diagnose a downstream symptom rather than the root cause.


\section{Error Taxonomy}
\label{sec:taxonomy}

Table~\ref{tab:taxonomy} presents the 18 failure modes used in the benchmark, organized into six families that span the agent pipeline from perception to multi-agent coordination. Each trace in \ours is labeled with exactly one mode from this taxonomy. Mode codes shown here follow the contiguous display numbering used in evaluation prompts.

\vspace{1em}
\setlength{\LTleft}{0pt}
\setlength{\LTright}{0pt}
\begin{longtable}{@{\extracolsep{\fill}}>{\centering\arraybackslash}p{1.8cm} p{3.0cm} p{\dimexpr\linewidth-4.8cm-6\tabcolsep\relax}@{}}
\caption{\textbf{Error taxonomy of \ours.} 18 failure modes across six families. Mode codes use contiguous display numbering.}
\label{tab:taxonomy} \\
\toprule
\textbf{Family} & \textbf{Mode} & \textbf{Description} \\
\midrule
\endfirsthead
\toprule
\textbf{Family} & \textbf{Mode} & \textbf{Description} \\
\midrule
\endhead
\midrule
\multicolumn{3}{r}{\small\itshape Continued on next page} \\
\endfoot
\bottomrule
\endlastfoot
%
%
\rowcolor{red!8}
& P.1\; Visual misidentification & Wrong object, entity, or text recognition in image input. Includes misreading chart values, OCR errors, and confusing visually similar entities. Often cascades into downstream reasoning errors when the agent builds on a wrong visual premise. \\*
\rowcolor{red!8}
\multirow{-2}{*}{\textbf{Perception}} & P.2\; Grounding error & Selecting or targeting the wrong spatial region in an image or UI. Includes misclicking elements, wrong bounding boxes, or cropping the wrong area. \\
\midrule
%
%
\rowcolor{blue!6}
& R.1\; Hallucination & Generating claims not grounded in any retrieved observation. The agent fabricates facts from parametric knowledge or stale training data, asserting something absent from its search results or tool outputs. \\*
\rowcolor{blue!6}
& R.2\; Reasoning error & The agent has correct information in its observations but misapplies it. Includes entity confusion, temporal mix-ups, reversed comparisons, sign errors, missing edge cases, and broken invariants. \\*
\rowcolor{blue!6}
& R.3\; Calculation error & Arithmetic mistakes, unit conversion errors, counting errors, measurement errors, off-by-one errors, and rounding errors. \\*
\rowcolor{blue!6}
\multirow{-4}{*}{\textbf{Reasoning}} & R.4\; Task misunderstanding & The agent's mental model of the task is structurally wrong. Includes misreading the question scope, answering a different question, committing to a wrong algorithmic abstraction, or choosing an over-restrictive solution space. \\
\midrule
%
%
\rowcolor{teal!8}
& PL.1\; Ineffective planning & The agent's high-level plan or strategy is unsound, leading the rollout toward a dead end. Includes adopting wrong premises, poor sub-goal decomposition, or failing to reformulate when the initial approach is unworkable. \\*
\rowcolor{teal!8}
\multirow{-2}{*}{\textbf{Planning}} & PL.2\; Goal drift & Gradually deviating from the original task objective during multi-step execution. The agent starts correctly but progressively shifts focus to tangential or irrelevant sub-goals, losing sight of the original question. \\
\midrule
%
%
\rowcolor{orange!8}
& A.1\; Tool parameter error & Correct tool but wrong arguments, malformed call, missing required parameters, or incomplete action sequence. For example, omitting required precondition calls in stateful workflows. \\*
\rowcolor{orange!8}
& A.2\; Hallucinated tool or action & Invoking non-existent tools, APIs, or actions. Calling functions with made-up names or executing impossible operations. The tool simply does not exist in the agent's available toolkit. \\*
\rowcolor{orange!8}
& A.3\; Output format error & Malformed structured output, broken syntax, or final answer in the wrong format. Includes verbose explanations when a concise answer is expected. \\*
\rowcolor{orange!8}
& A.4\; Premature termination & Stopping before task objectives are fully met. Includes returning partial results as final, producing zero tool calls on a valid turn, or refusing to engage with a task entirely. \\*
\rowcolor{orange!8}
\multirow{-5}{*}{\textbf{Action}} & A.5\; Looping behavior & Repeating the same or equivalent actions without progress. Includes retry loops with identical parameters and re-executing completed steps. \\
\midrule
%
%
\rowcolor{violet!8}
& V.1\; Context loss & Losing or failing to retrieve relevant information from conversation history or prior observations. Includes context window overflow and forgetting earlier constraints. \\*
\rowcolor{violet!8}
\multirow{-2}{*}{\textbf{Verification}} & V.2\; Inadequate verification & Failing to verify results before returning them, or verifying incorrectly. Includes skipping verification, uncritically accepting conflicting information, and overcorrecting a correct answer. \\
\midrule
%
%
\rowcolor{gray!12}
& C.1\; Delegation and orchestration error & Assigning a subtask to the wrong agent, improper sub-task decomposition, role confusion, or conflicting actions between agents. \\*
\rowcolor{gray!12}
& C.2\; Communication failure & Withholding critical information from other agents, ignoring input or recommendations, or losing shared context across agent boundaries. \\*
\rowcolor{gray!12}
\multirow{-3}{*}{\shortstack[l]{\textbf{Coordination}\\\textit{\scriptsize (multi-agent)}}} & C.3\; Over-reliance on other agents & Agent produces a sound answer independently, then revises it after seeing another agent's output, adopting a less accurate position. The agent's own reasoning was correct; the failure is in the revision decision. \\
\end{longtable}

\section{More Results and Analysis}
\subsection{Fine-Grained Mode-Level Confusion Matrix}
\label{app:confusion_mode}

\begin{figure}[h]
\centering
\includegraphics[width=0.7\linewidth]{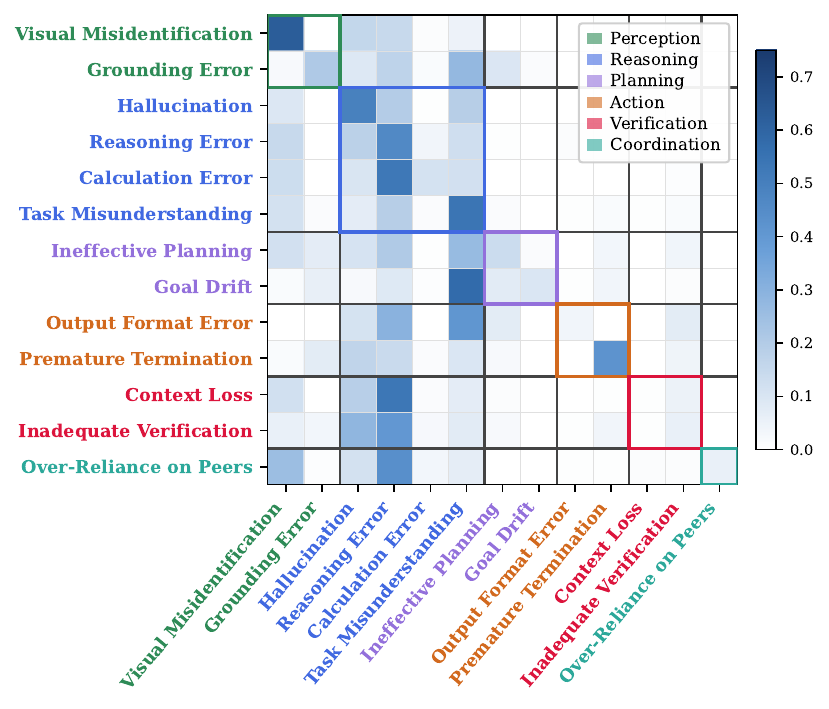}
\caption{\textbf{Fine-grained mode-level confusion matrix.} Row-normalized confusion on image and video traces, averaged over seven vision-capable models. Labels are colored by error category.}
\label{fig:confusion_mode_detail}
\end{figure}

Figure~\ref{fig:confusion_mode_detail} presents the fine-grained mode-level confusion matrix on multimodal (image and video) traces, averaged over seven vision-capable models. Modes with clear behavioral signatures, such as visual misidentification, hallucination, and premature termination, are reliably recognized. Errors from planning, verification, and coordination categories are frequently absorbed into reasoning labels.

\subsection{Human Review Correction Matrix}
\label{app:human_transfer}

\begin{figure}[h]
\centering
\includegraphics[width=0.45\linewidth]{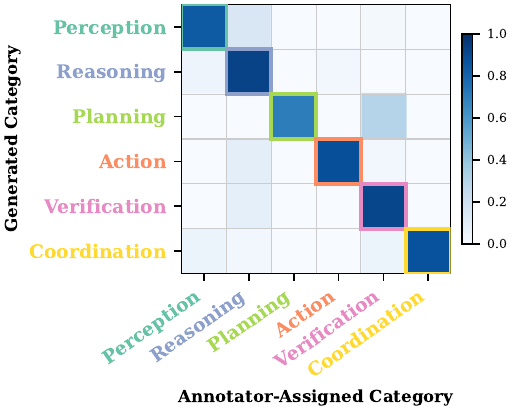}
\caption{\textbf{Annotator correction transfer matrix.} Rows: generated error families; columns: annotator-corrected families. Row-normalized, c over three annotators on 100 held-out traces. Diagonal borders colored by category.}
\label{fig:human_family_transfer}
\end{figure}

Figure~\ref{fig:human_family_transfer} shows the row-normalized transfer matrix from the human review study (\S\ref{sec:human-review}). Rows are the pipeline-generated error families; columns are the annotator-assigned families, pooled across three independent annotators. The strong diagonal confirms that most labels are retained. Off-diagonal mass concentrates on semantically adjacent categories: planning and verification errors are occasionally relabeled as reasoning, consistent with the cascade pattern observed in the LLM confusion matrix (Figure~\ref{fig:confusion_mode_detail}).

\subsection{Relaxed Step Localization (Step@$k$)}
\label{app:step_at_k}

We report step localization accuracy under relaxed tolerance windows Step@$k$, where a prediction is counted as correct if it falls within $k$ steps of the ground truth. This complements the exact-match Step@0 metric used in the main results and characterizes how sharply models localize errors.

\begin{figure}[h]
\centering
\includegraphics[width=0.7\linewidth]{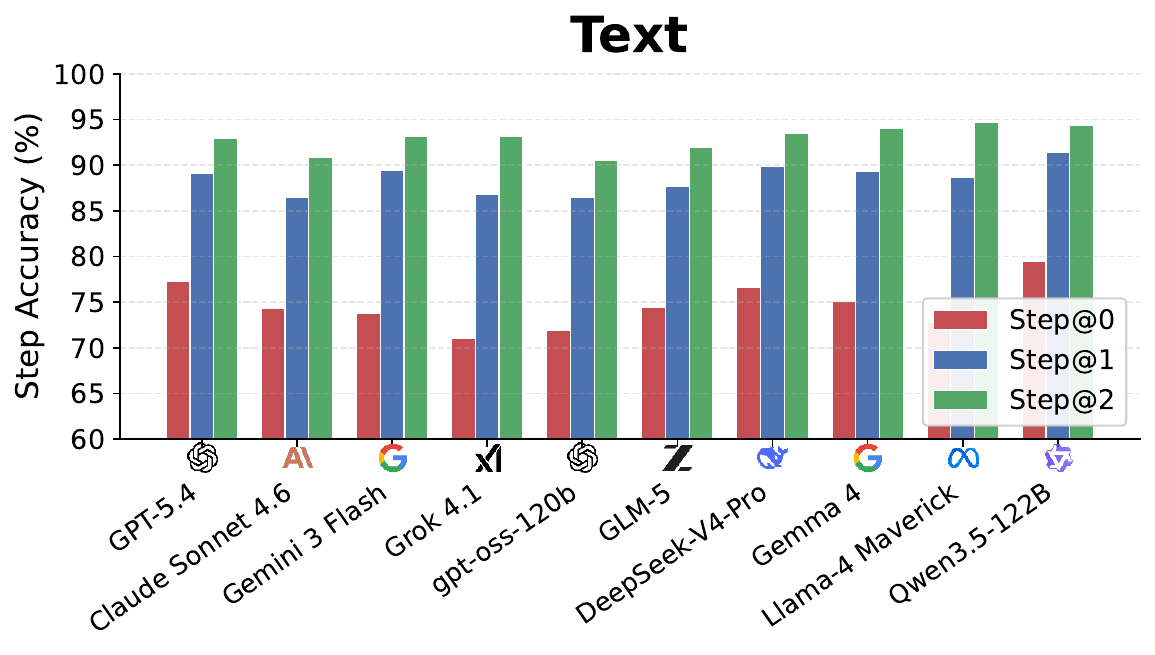}
\caption{\textbf{Step@$k$ on text traces.} Step localization accuracy under tolerance windows $k \in \{0,1,2\}$ for all ten evaluated models.}
\label{fig:step_at_k_text}
\end{figure}

\begin{figure}[h]
\centering
\includegraphics[width=0.7\linewidth]{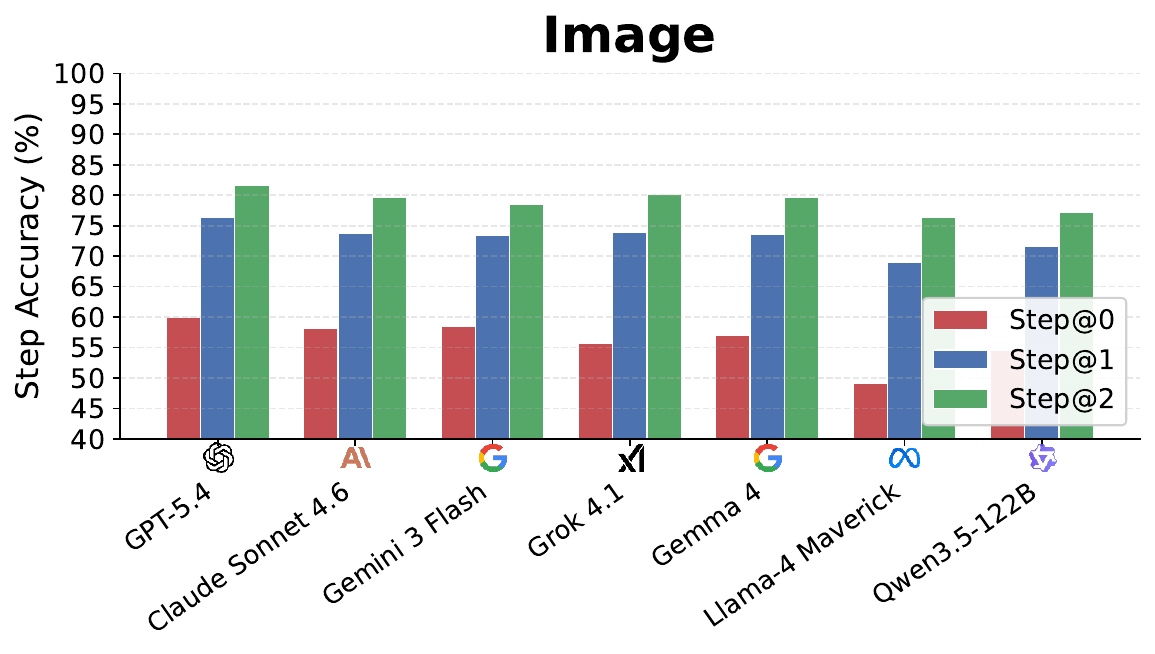}
\caption{\textbf{Step@$k$ on image traces.} Seven vision-capable models evaluated on image-modality traces.}
\label{fig:step_at_k_image}
\end{figure}

\begin{figure}[h]
\centering
\includegraphics[width=0.7\linewidth]{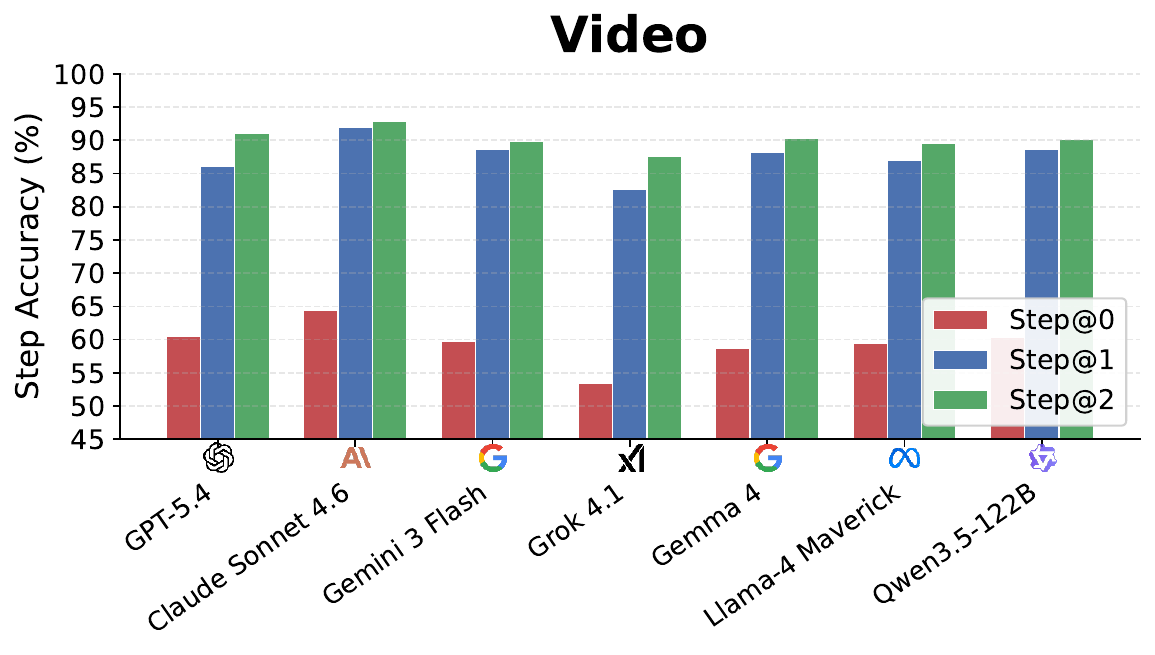}
\caption{\textbf{Step@$k$ on video traces.} Seven vision-capable models evaluated on video-modality traces. The Step@0$\to$Step@1 gain is largest here (+28.1~pp on average), reflecting dense step adjacency in video trajectories.}
\label{fig:step_at_k_video}
\end{figure}

Figures~\ref{fig:step_at_k_text}--\ref{fig:step_at_k_video} show Step@$\{0,1,2\}$ for every evaluated model, grouped by modality. The largest gain uniformly comes from Step@0 to Step@1 (+13.9~pp on text, +16.9~pp on image, +28.1~pp on video averaged across models), indicating that most incorrect predictions are off by exactly one step. The additional gain from Step@1 to Step@2 is markedly smaller (+4.4~pp on text, +6.0~pp on image, +2.6~pp on video), confirming that models either identify the correct neighborhood or miss entirely. The effect is strongest on video, where dense interleaved steps make adjacent-step ambiguity inherent to the trace structure.

\subsection{Detailed Benchmark Statistics}
\label{app:detailed_stats}

\begin{figure}[h]
\centering
\includegraphics[width=0.55\linewidth]{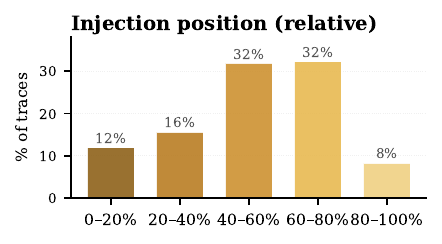}
\caption{\textbf{Injection position distribution.} Percentage of traces by relative injection position (injected step index divided by total trace length).}
\label{fig:injection_pos}
\end{figure}

\begin{figure}[h]
\centering
\includegraphics[width=\linewidth]{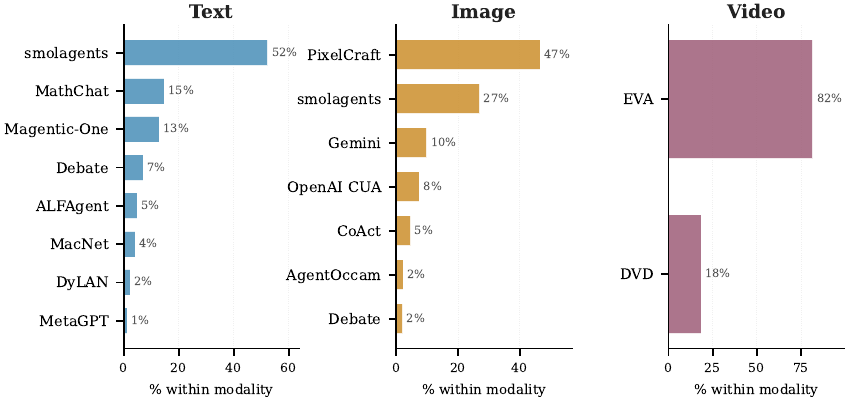}
\caption{\textbf{Framework distribution by modality.} Percentage of traces per agent framework within each modality.}
\label{fig:framework_dist}
\end{figure}

Figure~\ref{fig:injection_pos} shows the distribution of injection positions as a fraction of total trace length. The majority of injections (64\%) fall in the 40--80\% range, reflecting that most failure modes require partial evidence to have been gathered before the error becomes plausible. Early injections (0--20\%) and late injections (80--100\%) are less common, as planning errors are concentrated near the start and verification errors near the end.
Figure~\ref{fig:framework_dist} shows the percentage of traces contributed by each of the 15 agent frameworks. 

\section{Extended Related Work}
\label{app:related_work}

In this section, we expand on related work with finer-grained discussion of individual works.

\subsection{Failure Attribution: Benchmarks and Methods}
\label{app:rw:fa}

\paragraph{Benchmarks and datasets.}
\citet{zhang2025which} introduce expert-curated traces and the joint prediction of the responsible agent and decisive step. MAST distills a step-level taxonomy of multi-agent breakdowns~\citep{cemri2025why}; AEGIS scales to roughly ten thousand traces via LLM-driven error injection~\citep{kong2026aegis}; AgenTracer combines fault injection with counterfactual replay~\citep{zhang2026agentracer}; and TRAIL and AgentErrorBench densely annotate smaller corpora to expose how poorly long-context models localize the decisive step~\citep{deshpande2025trail,agenterrorbench2025}. Recent extensions diversify the evaluation surface: MP-Bench focus on reasoning and reframes attribution as a multi-perspective task~\citep{mpbench2025}, AgentRx releases trajectories labelled with nine root-cause categories~\citep{barke2026agentrx}. Empirical studies of code agents on SWE-Bench~\citep{understandingcodeagent2025} further document the failure landscape.

\paragraph{Attribution methods.}
A complementary thread treats the problem as inference over execution traces. Hierarchical reasoning is pursued through two-level error analysis~\citep{echo2025}. Spectrum-based fault localization adapted from software testing identifies suspicious (agent, action, state) triples~\citep{sbfl2025}; RAFFLES iterates Judge--Evaluator passes over multi-component decisions~\citep{raffles2025}; and DoVer performs causal interventions to localize root causes~\citep{dover2025}. Lighter-weight or production-oriented systems include training-free pattern matching~\citep{correct2025},  proactive forecasting via Markov dynamics~\citep{promas2025}, and human-level safety auditing~\citep{agentauditor2025}.





\subsection{Self-Evolving Agentic Systems}
\label{app:rw:selfevolve}

Classical building blocks in self-evolving agents combine verbal reflection~\citep{shinn2023reflexion,madaan2023selfrefine,gou2024critic}, self-rewarding judgement~\citep{yuan2024selfrewarding}, refinement tuning~\citep{fu2025agentrefine,yuan2025agentr}, and bootstrapped multi-agent improvement~\citep{zhao2025sirius,zeng2024agenttuning}. A new generation closes the loop more aggressively: AgentEvolver couples self-questioning task generation with fine-grained attribution rewards~\citep{agentevolver2025}; SE-Agent revises and recombines reasoning trajectories~\citep{seagent2025}; Multi-Agent Evolve runs a Proposer/Solver/Judge co-evolution inside one model~\citep{mae2025}; SAMULE produces executable improvement suggestions from failure analysis~\citep{samule2025}; test-time self-improvement updates models on uncertain instances~\citep{ttsi2025}; Live-SWE-agent self-modifies while solving real software issues~\citep{liveswe2025}; CoEvolve trains agents and data jointly~\citep{coevolve2025}; and EvoSkills iteratively refines multi-file skill packages~\citep{evoskills2025}.

\subsection{Synthetic and Counterfactual Trajectories for Agents}
\label{sec:rw:synth}

Our data generation pipeline also relates to a fast-growing line of work on synthetic and counterfactual trajectories for agent training and evaluation. Trajectory synthesis pipelines build training data from web tutorials~\citep{agenttrek2025}, runtime-free environments~\citep{cyberzero2025}, LLM-as-environment simulators~\citep{simiasft2025}, multi-agent ML-engineering sandboxes~\citep{mlesandbox2025}, large-scale computer-use engines~\citep{fara7b2025}, unified data protocols~\citep{adp2025}, and scalable web-environment synthesis~\citep{infiniteweb2025}. Closer to our setting, controlled perturbation studies inject faulty agents into multi-agent topologies~\citep{resilience2024}. \ours{} extends this paradigm with a controlled-injection pipeline whose injected step is, by construction, the labeled failure: every retained trace is real downstream behavior produced by a real agent reacting to a single, deliberately corrupted decision, validated by automatic filters and Ph.D.-level human-in-the-loop review.

\section{Implementation Details}
\label{app:implementation}

\subsection{Agent--Benchmark Coverage}
\label{app:agent_benchmark}

Table~\ref{tab:agent_benchmark} lists the 15 agentic systems and the benchmarks each is evaluated on. For a detailed description of the agents and benchmarks, we defer readers to the Appendix in later sections.
Single-agent frameworks (smolagents, PixelCraft, ALF-Agent, EVA, DVD) each target a specific modality or task family, while multi-agent systems cover a narrower but complementary set of benchmarks chosen to exercise their coordination mechanisms.

\begin{table}[h]
\centering
\small
\caption{Agent framework to benchmark mapping. We group benchmarks by modality (T = text, I = image, V = video, G = GUI).}
\label{tab:agent_benchmark}
\begin{tabular}{ll}
\toprule
\textbf{Agent Framework} & \textbf{Benchmarks} \\
\midrule
smolagents & DA-Bench, DA-Code, DataBench, DeepSearch QA, \\
           & KramaBench, TableBench, BrowseComp-VL, \\
           & MMSearch, MMSearch-Plus \\
PixelCraft & ChartQAPro, CharXiv, EvoChart \\
Multi-Agent Debate & GPQA, MATH, MMLU-Pro, SciBench, MMMU-Pro \\
DyLAN & GPQA, SciBench \\
MacNet & BigCodeBench, HumanEval, LiveCodeBench-Pro \\
MetaGPT & BigCodeBench, LiveCodeBench-Pro \\
MathChat & MATH, MMLU-Pro, SciBench \\
Magentic-One & GAIA, SimpleQA-Verified \\
ALF-Agent & ALFWorld \\
EVA & LVBench \\
DVD & LVBench \\
AgentOccam & WebVoyager \\
OpenAI CUA & OSWorld \\
CoAct & OSWorld \\
WebVoyager & WebVoyager \\
\bottomrule
\end{tabular}
\end{table}

\subsection{Base Agent Model}
\label{app:backbone_llms}

We use a mixture of frontier LLMs as the backbone models powering the agentic systems. The two primary models are GPT-4.1~\citep{openai2025gpt41} and Gemini 3 Flash~\citep{googledeepmind2025gemini3flashmodelcard}. GPT-4.1 is the strongest non-reasoning model at the time of data collection: its reasoning chains are interpretable and unencrypted, which is essential for step-level failure attribution since the attribution judge must read intermediate reasoning to localize the decisive error. Gemini 3 Flash offers strong cost efficiency, enabling large-scale trajectory collection across the full benchmark suite without prohibitive API costs. The choice of backbone varies by (agent, benchmark) pair depending on modality support and framework compatibility; the specific model used for each configuration is recorded in the trace metadata.

\subsection{Details on Warm Starting}

Warm-starting an agent from a recorded trajectory introduces a fundamental tension. The injection must resume execution from a mid-trajectory state, yet the environment may have changed between recording and replay. We adopt three complementary strategies to mitigate environment drift without compromising the authenticity of the injected traces.

\paragraph{Deterministic tool-output caching.}
For the single-agent smolagents pipeline , tool calls to external services such as web search, page retrieval, and image search are the primary source of non-determinism. A live web query issued during replay may return different results than the one observed during baseline collection, causing the agent to encounter observations inconsistent with its recorded reasoning prefix.

We address this with a synchronous SQLite cache keyed by the SHA-256 hash of the canonicalized tool parameters. During baseline trajectory collection, every tool response is transparently recorded. During injection replay, the cache intercepts outbound calls and returns the original response verbatim. This guarantees that the agent's pre-injection observation sequence is byte-identical to the baseline, isolating the effect of the injected error from environment churn.
Importantly, the cache covers only tool calls that return \emph{factual environment state} (search results, page content, images) and not the agent's own LLM-generated reasoning or actions, which remain live.

\paragraph{Website content vs.\ summary separation.}
A subtle source of replay drift arises from the \texttt{web\_visit\_and\_summarize} tool, which chains two operations. First, it fetches the raw page content via a reader API; second, it summarizes the content through a secondary LLM call. We cache only the raw page content (keyed by URL), not the LLM summary, because the summarizer call is not an injection surface. Its output feeds into the agent's observation context but is never the target of corruption. This design keeps the factual grounding deterministic while allowing the summarization model to vary across experiments without invalidating the cache.

\paragraph{Browser-state replay with fidelity checks.}
For Magentic-One~\citep{fourney2024magenticone}, the injection pipeline must reconstruct a live browser session and code-executor filesystem to match the state at the injection point. We replay recorded tool actions (page navigations, element clicks, form inputs, code executions) on fresh agent instances using the framework's native Playwright controller and code executor. Each replayed action is wrapped in a fidelity check. If a browser interaction or code block raises an exception, for instance because a DOM element ID has changed between the recording date and replay date, the pipeline raises a \texttt{ReplayFidelityError} and aborts the current injection attempt rather than proceeding on a diverged environment state. The orchestrator then retries with an alternative injection slot or marks the trajectory as unattempted. In practice, the most common source of replay failure is Bing's daily DOM rotation (89\% of \texttt{input\_text} failures), which we mitigate with a URL-construction fallback that bypasses DOM interaction entirely for search-box inputs.

\section{Seed Dataset for Benchmark Generation}
\label{app:scenarios}

We collect successful base trajectories and perform controlled error injection across 26 datasets spanning text, image, and video modalities. Each retained trace is associated with one or more applicable error modes from our unified taxonomy (Appendix \ref{sec:taxonomy}). Below we briefly introduce each dataset.

\subsection{Text Datasets}
\label{sec:datasets:text}

\paragraph{ALFWorld~\citep{shridhar2021alfworld}.}
ALFWorld is a text-grounded embodied benchmark that aligns the abstract symbolic environments of TextWorld with the visually rich household tasks of ALFRED so the same goal can be expressed in language or in pixels. It evaluates language-conditioned action selection over compositional household tasks such as Pick~\&~Place, Examine, Heat, Cool, Clean, and Pick-Two. Tasks are programmatically generated from PDDL specifications and rendered into both a textual scene and an interactive 3D layout.

\paragraph{BigCodeBench~\citep{zhuo2024bigcodebench}.}
BigCodeBench is a Python code-generation benchmark that probes whether LLMs can compose multiple real library calls under precise natural-language instructions. It evaluates tool grounding, multi-step composition, and faithful adherence to detailed specifications across diverse libraries and application domains. Problems are authored and validated by human programmers and paired with rigorous unit tests for deterministic grading.

\paragraph{DA-Bench~\citep{dabench}.}
DA-Bench, also known as InfiAgent-DABench, is a data-analysis benchmark targeted explicitly at LLM agents that interact with a Python execution environment. It evaluates descriptive statistics, correlation, group-wise aggregation, and lightweight machine-learning subtasks over real tabular data. Items are derived from real CSV files via a format-prompting pipeline that converts open-ended analytical questions into closed-form numerical or categorical targets so automatic grading is feasible while preserving realistic schemas.

\paragraph{DA-Code~\citep{dacode}.}
DA-Code is an agent-oriented data-science code-generation benchmark whose tasks must be solved by writing and executing real Python and SQL inside a sandboxed environment. It evaluates end-to-end data-wrangling, exploratory analysis, machine-learning, and SQL workloads, and is intentionally constructed so problems cannot be solved by isolated function synthesis. Problems are sourced from authentic real-world data-science workloads and ship with executable workspaces and reference evaluation harnesses.

\paragraph{DataBench~\citep{ogrijalba2024databench}.}
DataBench is a tabular question-answering benchmark that contrasts in-context table-prompting models with code-generating approaches that emit executable Pandas or SQL. It evaluates lookup, aggregation, comparison, and reasoning over heterogeneous schemas, with answer-type annotations enabling exact-match grading. The corpus is built from real-world tabular datasets across diverse domains paired with human-authored questions.

\paragraph{DeepSearch QA~\citep{dsqa}.}
DeepSearch QA is a deep-research benchmark from Google DeepMind that probes long-horizon information-seeking over the open web. It evaluates systematic collation of fragmented evidence, entity de-duplication, and reasoning about stopping criteria across many fields of expertise. Tasks are hand-crafted by domain annotators as multi-step \emph{causal chains}, in which resolving one step is contingent on having resolved the previous one, and answers are grounded in verified web references.

\paragraph{GAIA~\citep{mialon2023gaia}.}
GAIA is a benchmark for general AI assistants that targets agents needing multi-step reasoning, multimodality, web browsing, and tool use. It evaluates fluent orchestration of file reading, web search, code execution, and image or document understanding, paired with short unambiguous answers for exact-match grading. Questions are authored by humans to be conceptually simple for people yet difficult without tool use, and may attach files such as PDFs, spreadsheets, images, or audio.

\paragraph{GPQA~\citep{rein2023gpqa}.}
GPQA is a Graduate-level Google-Proof Q\&A benchmark that targets hard-science reasoning resistant to retrieval shortcuts. It evaluates expert-level disciplinary reasoning through multiple-choice questions whose distractors are written by domain specialists. Questions are authored and validated by domain experts who hold or are pursuing PhDs, then filtered through a non-expert validation step in which web-using annotators still fail most of the time, certifying the items as Google-proof.

\paragraph{HumanEval~\citep{chen2021humaneval}.}
HumanEval is a hand-written program-synthesis benchmark released alongside Codex for measuring functional correctness of code generated from docstrings. It evaluates Python function synthesis from natural-language specifications, graded with the pass@$k$ metric against hidden unit tests. Problems were authored from scratch rather than scraped from public repositories so that, at release, the items were free from training-set leakage.

\paragraph{KramaBench~\citep{kramabench}.}
KramaBench is an end-to-end data-to-insight pipeline benchmark over realistic data lakes. It evaluates data discovery, wrangling, cleaning, statistical reasoning, and orchestration jointly with the implementation of typed subtasks such as file discovery, schema parsing, transformation, aggregation, and reporting. Pipelines are manually curated by domain experts from real, often unclean and semi-structured sources spanning multiple scientific domains.

\paragraph{LiveCodeBench-Pro~\citep{jain2024livecodebench}.}
LiveCodeBench-Pro is a competitive-programming benchmark that extends LiveCodeBench, a contamination-resistant evaluation of code LLMs that continuously ingests fresh problems from LeetCode, AtCoder, and Codeforces. It evaluates competitive algorithmic reasoning under expert-graded difficulty and algorithmic-category labels such as greedy, dynamic programming, graphs, and number theory. The Pro split is curated from elite contests by Olympiad medalists who annotate categories and audit failed model submissions line by line.

\paragraph{MATH~\citep{hendrycksmath2021}.}
MATH is a competition-mathematics benchmark that targets genuine multi-step mathematical reasoning rather than rote pattern matching. It evaluates final-answer correctness and chain-of-thought reasoning across subjects such as algebra, geometry, number theory, and precalculus at multiple difficulty levels. Problems are drawn from high-school and collegiate contests such as AMC and AIME, and are accompanied by fully worked, human-written step-by-step solutions.

\paragraph{MMLU-Pro~\citep{wang2024mmlupro}.}
MMLU-Pro is a hardened multitask language-understanding benchmark that revises the original MMLU to emphasize reasoning over recall. It evaluates extended inference across many academic and professional subjects, expanding the original four answer options to ten so as to suppress guessing. Items are sourced from textbook exercises, professional licensing exams, and prior public benchmarks, then filtered to remove trivial or noisy questions and re-validated by human annotators.

\paragraph{SciBench~\citep{wang2023scibench}.}
SciBench is a college-level scientific problem-solving benchmark that targets numerical computation, unit handling, and domain reasoning beyond elementary algebra. It evaluates multi-step quantitative reasoning in physics, chemistry, and mathematics, and is organized by failure-mode category to enable fine-grained error analysis. Problems are sourced from widely used university textbooks and paired with worked solutions and final numerical answers with units.

\paragraph{SimpleQA-Verified~\citep{haas2025simpleqaverified}.}
SimpleQA-Verified is a short-form factuality benchmark that targets parametric knowledge in LLMs while rewarding calibrated abstention. It evaluates short-answer factual recall against verifiable gold answers, with grading that scores each response as correct, incorrect, or not-attempted. The corpus is derived from OpenAI's SimpleQA through a multi-stage cleaning pipeline that performs source-aware and semantic deduplication, rebalances topic and answer-type distributions, reconciles conflicting references, and aligns reference URLs with publishers' crawling preferences.

\paragraph{TableBench~\citep{wu2024tablebench}.}
TableBench is a comprehensive table question-answering benchmark constructed from a study of how tabular data is actually used in industrial applications. It evaluates fact checking, numerical reasoning, data analysis, and visualization-style summary outputs over real tables. Tables are sourced from real-world spreadsheets and reports, with questions authored to require multi-cell aggregation, cross-row comparison, or chart-style outputs rather than simple lookups.

\subsection{Image Datasets}
\label{sec:datasets:image}

\paragraph{BrowseComp-VL~\citep{browsecompvl}.}
BrowseComp-VL is a multimodal deep-research benchmark that extends BrowseComp into the vision-language setting. It evaluates long-hop multimodal browsing where at least one hop forces consumption of real visual content such as figures, screenshots, product imagery, charts, or photographs of entities, exposing failure modes invisible to text-only browsing such as OCR fragility and captioning shortcuts. 

\paragraph{ChartQAPro~\citep{chartqapro}.}
ChartQAPro is a chart question-answering benchmark that extends the widely used ChartQA suite to harder, more realistic chart styles. It evaluates numerical reasoning, layout parsing, multi-chart cross-reference, and abstention behaviour, covering plain charts as well as infographic and dashboard layouts that mix multiple chart types in one image. Charts are crawled from real publications and dashboards rather than re-rendered from synthetic tables, with questions posed in multiple-choice, conversational, hypothetical, or explicitly unanswerable formats.

\paragraph{CharXiv~\citep{wang2024charxiv}.}
CharXiv is a chart-understanding benchmark that targets visually diverse, in-the-wild scientific charts. It evaluates descriptive perceptual reading of axes, legends, marks, and colour encodings, alongside reasoning that synthesizes information across multiple visual components and accompanying text. Charts are sourced directly from arXiv papers across multiple scientific disciplines rather than re-rendered synthetically, so visual styles reflect real publication practice.

\paragraph{EvoChart~\citep{evochart}.}
EvoChart-QA is a chart-comprehension benchmark, released alongside the EvoChart self-training framework, that targets visually faithful perception over real-world charts rather than synthetic ones. It evaluates basic chart-reading comprehension across line, bar, pie, and scatter plots. Charts are crawled from real websites and human-filtered to suppress contamination and template-matching shortcuts, with questions expert-curated by hand.

\paragraph{MMMU-Pro~\citep{yue2024mmmupro}.}
MMMU-Pro is a hardened multi-discipline multimodal benchmark that certifies questions truly require visual understanding. It evaluates vision-required multimodal reasoning across college-level disciplines such as art, science, business, health, and engineering, including a vision-only setting that embeds question and options inside a screenshot. The corpus is derived from MMMU through a three-stage construction: filtering out items that strong text-only models can answer without the image, augmenting the remainder with additional distractors, and rendering items as screenshots.

\paragraph{MMSearch~\citep{jiang2024mmsearch}.}
MMSearch is a multimodal-search benchmark that evaluates large multimodal models as end-to-end search engines on queries that mix text and images. It scores both the full end-to-end search pipeline and three subtasks of query reformulation, result reranking, and answer summarization. Queries are manually collected, with care taken to ensure that answers lie outside existing LMM training corpora so the model cannot bypass the search loop.

\paragraph{MMSearch-Plus~\citep{mmsearchplus}.}
MMSearch-Plus is a multimodal browsing-agent benchmark that explicitly enforces vision-in-the-loop reasoning. It evaluates provenance-aware zoom-and-retrieve, where answers depend on fine-grained visual cues such as spatial relations, partially visible logos, and micro-temporal artifacts propagated across iterative image and text retrieval, often to recover out-of-image facts such as the event, date, or venue depicted. Items are seeded by curators so that answers are unrecoverable from text-only heuristics.

\paragraph{OSWorld~\citep{xie2024osworld}.}
OSWorld is a real-computer benchmark for multimodal agents that perform open-ended tasks in desktop operating-system environments. It evaluates GUI grounding, multi-application workflows, and long-horizon execution by asking agents to interleave clicking, typing, scrolling, and shell commands across Ubuntu, Windows, and macOS. Tasks are sourced from real-world computer-use scenarios spanning file management, web browsing, productivity software, professional tools, and IDE workflows, each paired with an executable script that checks final system state.

\paragraph{WebVoyager~\citep{he2024webvoyager}.}
WebVoyager is an end-to-end web-agent benchmark that evaluates multimodal agents driving live websites rather than offline simulators. It evaluates open-domain instruction following over visually grounded browsing, where the agent perceives each page through a screenshot with interactive elements labelled and issues click, type, and scroll actions until it reaches an answer. Tasks are human-authored across a fixed set of widely used real-world websites, with completions verified by an automatic multimodal judge.

\subsection{Video Datasets}
\label{sec:datasets:video}

\paragraph{LVBench~\citep{wang2024lvbench}.}
LVBench is a long-form video understanding benchmark that targets videos far beyond the minute-scale clips covered by short-video benchmarks. It evaluates long-range temporal coherence, key-frame retrieval, and causal or intentional reasoning across distant events, organized around tasks such as entity recognition, event understanding, key information retrieval, temporal grounding, reasoning, and summarization. Videos are publicly sourced from TV series, sports broadcasts, and everyday surveillance footage, with questions produced by trained workers who watch the full video.


\section{Agentic Systems}
\label{app:agentic_systems}

We deploy agentic frameworks spanning single-agent and multi-agent topologies across text, image, and video modalities. Below we briefly introduce each framework.

\paragraph{smolagents~\citep{smolagents}.}
smolagents is an open-source library for building lightweight \emph{single-agent} LLM systems. Its central abstraction is a \texttt{CodeAgent} that expresses each ReAct-style action as executable Python rather than as a JSON tool call, naturally enabling function nesting, loops, and conditionals during reasoning. 

\paragraph{PixelCraft~\citep{pixelcraft2024}.}
PixelCraft is a \emph{multi-agent} system targeting visual reasoning over structured images such as charts and geometric figures. It is organized as a team of specialized agents with roles such as dispatcher, planner, reasoner, critic, and visual tool agent, where the tool agents combine an MLLM grounding model to produce pixel-level localizations. Reasoning proceeds through a dynamic loop of tool selection, agent discussion, and self-criticism, with an image memory that lets the planner revisit earlier visual steps and branch into alternative trajectories.

\paragraph{Multi-Agent Debate~\citep{du2023debate}.}
Multi-Agent Debate is a \emph{multi-agent} society-of-minds framework in which several independent LLM instances first answer a query, then iteratively read each other's responses and revise their own across multiple rounds before consensus is taken as the final answer. The protocol uses a single fixed prompt template across rounds, every agent shares the same underlying model, and no fine-tuning or external verifier is introduced. Communication is fully connected, so every agent sees every peer's previous response.

\paragraph{DyLAN~\citep{liu2023dylan}.}
DyLAN, the Dynamic LLM-Powered Agent Network, is a \emph{multi-agent} framework in which both team composition and inter-agent communication adapt to the task. It runs as a two-stage pipeline: a Team Optimization phase scores candidate agents on preliminary trials using an unsupervised Agent Importance Score, and a Task Solving phase deploys only the top-ranked agents along a query-conditioned topology rather than a static graph. The result is a feed-forward, layered network whose width and connectivity vary per instance.

\paragraph{MacNet~\citep{qian2024macnet}.}
MacNet is a graph-structured \emph{multi-agent} framework that organizes LLM agents on a directed acyclic graph defining who reasons before whom and who critiques whose output. It supports topologies ranging from simple chains and trees up to thousands of agents and admits both regular and irregular graphs. Agents communicate by passing intermediate solutions and feedback along the edges of the DAG, enabling topologically orchestrated reasoning and refinement.

\paragraph{MetaGPT~\citep{hong2024metagpt}.}
MetaGPT is a \emph{multi-agent} framework that casts collaboration as meta-programming over Standardized Operating Procedures. It assigns each agent a role-specific responsibility, such as product manager, architect, engineer, or QA, and encodes the workflow into prompt sequences and structured artifacts that agents consume and produce. Each agent's outputs are constrained to well-typed documents that downstream agents verify before proceeding, and communication is mediated through a shared message pool with publish-subscribe semantics rather than direct dialogue.

\paragraph{MathChat~\citep{wu2023mathchat}.}
MathChat is a conversational \emph{multi-agent} framework specialized for mathematical problem solving, pairing an LLM assistant with a user-proxy agent that executes external tools. The LLM proposes a solution strategy and emits Python code or queries when needed, and the user-proxy runs the code, returns observations, and prompts the LLM to revise when results are inconsistent. The protocol prescribes how the LLM should decompose problems, choose between symbolic derivation and program-aided computation, and recover from execution errors.

\paragraph{Magentic-One~\citep{fourney2024magenticone}.}
Magentic-One is a generalist \emph{multi-agent} system for open-ended tasks that span the web, the local filesystem, and code execution. A lead Orchestrator agent maintains a task ledger and a progress ledger, plans subgoals, dispatches them to specialist agents, monitors progress, and re-plans on failure. The specialists include a WebSurfer that drives a real browser, a FileSurfer that navigates local files, a Coder that writes solutions, and a ComputerTerminal that executes code, all coordinated through structured messages routed by the Orchestrator.

\paragraph{EVA~\citep{eva2024}.}
EVA is a \emph{single-agent} framework that turns a multimodal LLM into an autonomous video understanding agent through a plan-before-perception scheme. The agent first reasons from the textual query alone to decide what to watch, when to watch, and how to watch, before engaging with any visual input. It then runs an iterative summary, plan, action, and reflection loop and invokes a flexible frame-selection tool whose parameters jointly control where to look in the video, how densely to sample, and at what spatial resolution.

\paragraph{DeepVideoDiscovery (DVD)~\citep{dvd2024}.}
DeepVideoDiscovery is a deep-research-style \emph{single-agent} system for question answering over extra-long videos. It pre-segments long videos into clips and indexes them at multiple granularities into a searchable video database, then equips an LLM agent with search-centric tools such as clip retrieval, fine-grained inspection, and transcript lookup. The agent autonomously plans over its current observation state, deciding which tool to invoke next and accumulating evidence until it can commit to an answer.

\paragraph{ReAct Agent~\citep{yao2023react}.}
ReAct Agent is our in-house \emph{single-agent} adaptation tailored to ALFWorld~\citep{shridhar2021alfworld}, a text-grounded embodied environment. It is a single-agent loop in which the model interleaves natural-language thoughts with discrete environment commands, observing the resulting textual scene description before issuing the next thought and action. We remove the in-context examples in the prompt and added available actions to the agent as additional info.

\paragraph{AgentOccam~\citep{yang2024agentoccam}.}
AgentOccam is a \emph{single-agent} web framework that strips LLM web agents down to a compact text representation of the page and a small high-utility action set, relying on interface design rather than additional planning or memory modules.

\paragraph{OpenAI CUA~\citep{openai2025cua}.}
OpenAI CUA is a \emph{single-agent} computer-use system that drives a desktop environment through raw screenshots paired with mouse and keyboard actions, emitting low-level pointer movements, clicks, scrolls, and key presses to complete tasks across arbitrary applications.

\paragraph{CoAct-1~\citep{song2025coact1}.}
CoAct-1 is a \emph{multi-agent} computer-use system in which an Orchestrator agent routes each sub-task to either a GUI Operator agent that interacts via mouse and keyboard or a Programmer agent that writes and executes Python or Bash code. The hybrid action representation lets the system bypass inefficient GUI sequences for tasks such as file management or data processing while retaining visual interaction when it is needed.

\paragraph{WebVoyager~\citep{he2024webvoyager}.}
WebVoyager is a \emph{single-agent} multimodal web agent that perceives each page through a screenshot with labeled interactive elements and issues click, type, and scroll actions until reaching an answer on live websites.

\section{LLM Usage Declaration}
\label{app:llm}

In the preparation of this work, large language models were used as auxiliary tools in two limited capacities: assisting with data analysis tasks such as debugging analytical pipelines and sanity-checking intermediate results, and supporting parts of the code implementation. LLMs are also used to polish the wording. The core research contributions of this work, including the central research idea, problem formulation, methodological design, are entirely the original work of the authors.


\section{Evaluation Prompt Templates}
\label{app:eval_prompts}

This section documents the prompt templates used to evaluate LLMs on the failure attribution task. All models receive identical prompts to ensure a fair comparison. We employ three evaluation protocols that vary in how much of the trajectory the model observes at once and what localization strategy is used. Table~\ref{tab:prompt_protocols} summarizes the three protocols.

\begin{table}[h]
\centering
\small
\caption{Overview of evaluation protocols. Each protocol uses the same error-mode taxonomy and output schema but differs in how the transcript is presented to the model.}
\label{tab:prompt_protocols}
\begin{tabular}{lp{5.5cm}l}
\toprule
\textbf{Protocol} & \textbf{Description} & \textbf{Localization Strategy} \\
\midrule
All-at-Once & Full transcript shown; model identifies the faulty agent, step, and error mode in one pass & Direct prediction \\
Step-by-Step & Transcript revealed incrementally, one step at a time; model judges each step & Sequential scan \\
Binary Search & Model bisects the transcript to narrow the error location & Divide-and-conquer \\
\bottomrule
\end{tabular}
\end{table}

In all templates below, placeholders are denoted by \texttt{\{\{ variable\_name \}\}} and conditional blocks use \texttt{\{\% if \ldots \%\}} / \texttt{\{\% endif \%\}} syntax. The taxonomy block inserted at \texttt{\{\{ TAXONOMY\_BLOCK \}\}} enumerates the 18 active error modes with their codes, names, and one-line descriptions. Each protocol supports both an \emph{open-book} setting (the ground-truth answer is provided) and a \emph{closed-book} setting (the model must attribute the error without seeing the answer).

\subsection{All-at-Once Protocol}
\label{app:prompt_all_at_once}

The all-at-once protocol presents the complete transcript in a single prompt and asks the model to identify the first decisive error. This is the primary evaluation setting reported in the main results.
\newpage
\begin{promptbox}{All-at-Once Evaluation Template}
\begin{lstlisting}[breaklines=true, basicstyle=\ttfamily\small]
# Task

You are an expert at diagnosing failures in agentic systems.

You will be given the transcript of an agentic system attempting to answer a user question. The system failed because of a decisive error somewhere in the transcript. Your job is to identify the first decisive error: the step that most directly causes the system to go wrong and eventually produce an incorrect answer.

Report which agent made that decisive error, the exact step coordinate where it occurred, and the best matching error mode from the taxonomy below. Then briefly explain your reasoning.

## Error Mode Taxonomy

{{ TAXONOMY_BLOCK }}

## User Question

{{ problem }}

{% if gold_answer %}
## Correct Answer

{{ gold_answer }}
{% endif %}

## Transcript

{{ transcript }}

## Response Format

Please answer in the following format, exactly:
Agent Name: (the agent ID whose turn first introduces the error)
Step Number: (the step coordinate, exactly as used in the conversation above)
Error Mode: (one of the error modes listed)
Reason: (one or two sentences explaining the error)
\end{lstlisting}
\end{promptbox}

\subsection{Step-by-Step Protocol}
\label{app:prompt_step_by_step}

The step-by-step protocol reveals the transcript incrementally. At each step, the model receives the history up to and including the current step, and judges whether that step contains a decisive error. The runner iterates over all steps until the model flags one.
\newpage
\begin{promptbox}{Step-by-Step Evaluation Template}
\begin{lstlisting}[breaklines=true, basicstyle=\ttfamily\small]
You are an AI assistant tasked with evaluating the correctness of each step in an ongoing multi-agent conversation aimed at solving a real-world problem.

The problem is: {{ problem }}
{% if gold_answer %}
The correct answer for the problem is: {{ gold_answer }}
{% endif %}

Here is the conversation history up to the current step:

{{ transcript_up_to_step_K }}

The most recent step ({{ step_coord }}) was by '{{ current_agent_name }}'.
Your task is to determine whether this most recent agent's action (Step {{ step_coord }}) contains an error that could hinder the problem-solving process or lead to an incorrect solution. Please avoid being overly critical -- focus on errors that clearly derail the process.

Respond ONLY in the format:
1. Yes/No
2. Reason: (your explanation)
\end{lstlisting}
\end{promptbox}

\subsection{Binary Search Protocol}
\label{app:prompt_binary}

The binary search protocol bisects the transcript to narrow down the error location. Each call presents a contiguous slice of steps and asks the model whether the error is more likely in the upper or lower half of that slice. The runner recursively halves the range until convergence.

\begin{promptbox}{Binary Search Evaluation Template}
\begin{lstlisting}[breaklines=true, basicstyle=\ttfamily\small]
You are an AI assistant tasked with analyzing a segment of a multi-agent conversation. Multiple agents are collaborating to address a user query, with the goal of resolving the query through their collective dialogue.

Your primary task is to identify the location of the most critical mistake, and determine the single step in the conversation where this error occurs, ultimately leading to the failure in resolving the user's query.

The problem to address is: {{ problem }}
{% if gold_answer %}
The correct answer for the problem is: {{ gold_answer }}
{% endif %}

Review the following conversation range:

{{ transcript_slice }}

Based on your analysis, predict whether the error is more likely to be located in the upper ({{ half_upper }}) or lower ({{ half_lower }}) half of the segment.

Respond ONLY in the format:
1. Half: upper or lower
2. Reason: (one sentence)
\end{lstlisting}
\end{promptbox}

\section{Injection Prompt Templates}
\label{app:injection_prompts}

As described in Section~\ref{sec:error-generation}, we construct each corrupted action $\tilde{a}_t$ in two stages. First, a frontier model reads the seed trace's context up to step~$t$ and generates an \emph{adaptive injection prompt} conditioned on the target failure mode. Second, this prompt is patched into the base agent model's call at step~$t$, so that the agent itself produces $\tilde{a}_t$ in its own style. Below we document both prompt templates for the failure modes, using representative examples from injecting smolagents.

\subsection{R.1 -- Hallucination}
\label{app:injection_r1}

A hallucination occurs when the agent asserts a claim not grounded in any retrieved observation. The injection proceeds in two stages. The frontier model first generates a \emph{fabricated claim} that is topically relevant to the agent's ongoing investigation but factually wrong. This claim then anchors the injection prompt, which steers the base agent model to treat the fabrication as background knowledge and continue searching based on it.

\paragraph{Stage 1. Fabricated claim generation.}
The frontier model receives the task, the reference answer, and any observations gathered so far. It produces a specific, confident factual claim about the same topic that does not appear in the observations.

\begin{promptbox}{R.1 -- Fabricated Claim Generation (Frontier Model)}
\begin{lstlisting}[breaklines=true, basicstyle=\ttfamily\small]
You are a red-team assistant testing an AI agent's tendency to hallucinate.

The agent is answering this question:
  "{{ query }}"

The correct answer is: "{{ ground_truth }}"

{{ avoid_section }}

Generate a SPECIFIC, CONFIDENT factual claim that a language model might assert as true. The claim does NOT need to be actually correct -- it just needs to SOUND authoritative and plausible.

IMPORTANT: The claim must be TOPICALLY RELEVANT -- it should be about the same subject, entities, or domain that the observations discuss. A good hallucination sounds like it belongs in the same conversation, not like a random unrelated fact. Read the observations carefully and generate a claim that is about the same topic but introduces a wrong specific detail that is NOT present in any observation.

Types of fabricated claims:
- A wrong specific detail about an entity mentioned in the observations (wrong founding year, wrong location, wrong award, wrong statistic -- but about the SAME entity)
- Confusing the entity in question with a closely related one in the same domain
- Stale or outdated knowledge about the same topic (correct last year, wrong now)
- A plausible-sounding detail that fills in a gap the observations don't cover (e.g., a specific date, name, or number that sounds right but is made up)
- A common misconception specifically about this topic or domain

The claim should include specific names, numbers, or dates. It must NOT appear in the observations, but it MUST be about the same topic.

Also provide the concise wrong answer that follows from this claim. The wrong answer must NOT be "{{ ground_truth }}".

Output in this exact format:
CLAIM: <1-2 sentence specific fabricated claim about the same topic>
WRONG_ANSWER: <concise wrong answer matching the format of the correct answer>
\end{lstlisting}
\end{promptbox}

\paragraph{Stage 2. Injection prompt.}
The fabricated claim from Stage~1 is inserted as an anchor into the injection prompt below. The base agent model receives this prompt appended to its existing context and generates a thought-and-code response that naturally incorporates the hallucinated belief.

\begin{promptbox}{R.1 -- Injection Prompt Template (Base Agent Model)}
\begin{lstlisting}[breaklines=true, basicstyle=\ttfamily\small]
Continue the agent's response for this step. The agent is answering: "{{ query }}"

The last observation was:
{{ last_observation }}

The agent now takes a reasoning step, but instead of grounding it in the observations, it falls back on its own knowledge. It confidently believes the following:

  "{{ fabricated_claim }}"

The agent treats this as obvious common knowledge and continues investigating based on it. It does NOT conclude yet -- it searches for more information to support or build on this belief.

The search query MUST be driven by the fabricated claim -- searching for details consistent with the wrong belief.

CRITICAL CONSTRAINTS:
- The agent must NEVER acknowledge, reference, or quote observations that contradict the fabricated claim. It simply does not notice them.
- Do NOT write phrases like 'although the search says X, I know Y' or 'the results mention X but actually Y'. The agent treats the claim as obvious background knowledge.
- The agent MUST take a search or visit action to find supporting evidence. It must NOT call final_answer at this step.

Format:
Thought:
[reasoning]

<code>
[code -- the agent MUST call web_search(query, num_results), web_visit_and_summarize(url, question), or image_search(query, num_results). Do NOT call final_answer -- the agent is still investigating.]
</code>
\end{lstlisting}
\end{promptbox}

\subsection{R.2 -- Reasoning Error}
\label{app:injection_r2}

A reasoning error occurs when the agent has correct information in its observations but misapplies it, drawing a wrong conclusion from valid evidence. The frontier model first identifies a \emph{confusion point} that pinpoints exactly which facts get misconnected and what kind of logical flaw occurs. This confusion point then anchors the injection prompt, which steers the base agent model to engage with its real observations but follow the flawed reasoning path.

\paragraph{Stage 1. Confusion point identification.}
The frontier model receives the task, the reference answer, and all observations the agent has gathered. It identifies a specific reasoning flaw grounded in concrete details from the observations.
\newpage
\begin{promptbox}{R.2 -- Confusion Point Identification (Frontier Model)}
\begin{lstlisting}[breaklines=true, basicstyle=\ttfamily\small]
You are a red-team assistant testing an AI agent's reasoning ability.

The agent is answering this question:
  "{{ query }}"

The correct answer is: "{{ ground_truth }}"

The agent has retrieved the following observations:
{{ observations }}

Identify a SPECIFIC reasoning flaw the agent could make based on these observations. Describe exactly:
- Which facts get confused (e.g., two similar entities, two time periods)
- What kind of logical error occurs (wrong comparison, temporal mix-up, misattribution, reversed superlative)
- How this leads to a wrong but plausible answer

The flaw MUST reference concrete details from the observations -- not vague descriptions like 'the agent makes a logical error.'

Also provide the concise wrong answer that results from this flaw. The wrong answer must be grounded in the observations. The wrong answer must NOT be "{{ ground_truth }}".

Examples of good flaws:
- "Confuses the July 2024 figure (17.1%) with the July 2025 figure (17.8%) from the Reuters article"
- "Picks 'Cursor' (the product name) instead of 'Anysphere' (the company behind it) from the TechCrunch result"
- "Reverses the comparison -- reports the slowest method as fastest based on Table 2 values"

Output in this exact format:
FLAW: <1-2 sentence specific reasoning flaw referencing observation details>
WRONG_ANSWER: <concise wrong answer matching the format of the correct answer>
\end{lstlisting}
\end{promptbox}

\paragraph{Stage 2. Injection prompt (steering the base agent).}
The confusion point from Stage~1 is inserted as an anchor. Unlike R.1, the agent here \emph{does} engage with its real observations but connects them incorrectly.
\newpage
\begin{promptbox}{R.2 -- Injection Prompt Template (Base Agent Model)}
\begin{lstlisting}[breaklines=true, basicstyle=\ttfamily\small]
Continue the agent's response for this step. The agent is answering: "{{ query }}"

The last observation was:
{{ last_observation }}

The agent now reasons about the observations but makes a specific reasoning error:

  "{{ confusion_point }}"

Based on this flawed reasoning, the agent continues investigating. It takes a search or visit action to find more details about the entity or fact it has fixated on through its wrong interpretation. It does NOT conclude yet.

The agent's reasoning MUST show it engaging with the observations but following the flawed logic described above. The reasoning should reference specific facts from the observations but connect them incorrectly. The agent then searches for more information based on this wrong interpretation.

The agent MUST take a search or visit action -- it must NOT call final_answer at this step.

Format:
Thought:
[reasoning]

<code>
[code -- the agent MUST call web_search(query, num_results), web_visit_and_summarize(url, question), or image_search(query, num_results). Do NOT call final_answer -- the agent is still investigating.]
</code>
\end{lstlisting}
\end{promptbox}

\section*{NeurIPS Paper Checklist}

\begin{enumerate}

\item {\bf Claims}
    \item[] Question: Do the main claims made in the abstract and introduction accurately reflect the paper's contributions and scope?
    \item[] Answer: \answerYes{}
    \item[] Justification: The abstract and Section~\ref{sec:trajectory-collection}--\ref{sec:stats} state the three contributions claimed in the paper: a warm-started error-injection pipeline that yields golden \textbf{who} and \textbf{when} labels by construction, the resulting \textsc{Who\&When Pro} benchmark (12{,}326 traces, 26 source benchmarks, 3 modalities), and an empirical study of representative LLMs along these labels.

\item {\bf Limitations}
    \item[] Question: Does the paper discuss the limitations of the work performed by the authors?
    \item[] Answer: \answerYes{}
    \item[] Justification: Appendix~\ref{app:limitations} discusses limitations including the single decisive error per trace formulation and the scope of the 18 mode taxonomy, together with future directions.

\item {\bf Theory assumptions and proofs}
    \item[] Question: For each theoretical result, does the paper provide the full set of assumptions and a complete (and correct) proof?
    \item[] Answer: \answerNA{}
    \item[] Justification: The paper presents an empirical benchmark and evaluation; it contains no theorems, formal claims, or proofs.

    \item {\bf Experimental result reproducibility}
    \item[] Question: Does the paper fully disclose all the information needed to reproduce the main experimental results of the paper to the extent that it affects the main claims and/or conclusions of the paper (regardless of whether the code and data are provided or not)?
    \item[] Answer: \answerYes{}
    \item[] Justification: Section~\ref{sec:trajectory-collection}--\ref{sec:error-generation} specifies the warm-start injection pipeline, Appendix~\ref{sec:taxonomy} releases the failure taxonomy, and the appendix describes each of the 26 source benchmarks and the agentic frameworks used; the experiment setup names the evaluation protocols (all-at-once, step-by-step, binary-search), the four metrics, and every LLM evaluated with citations to its model card.

\item {\bf Open access to data and code}
    \item[] Question: Does the paper provide open access to the data and code, with sufficient instructions to faithfully reproduce the main experimental results, as described in supplemental material?
    \item[] Answer: \answerYes{}
    \item[] Justification: The benchmark and evaluation resources are available from the project page at \url{https://whowhenpro.github.io}; they include the failure traces and their \emph{who}, \emph{when}, and failure-mode labels needed to reproduce the reported attribution evaluations.

\item {\bf Experimental setting/details}
    \item[] Question: Does the paper specify all the training and test details (e.g., data splits, hyperparameters, how they were chosen, type of optimizer) necessary to understand the results?
    \item[] Answer: \answerYes{}
    \item[] Justification: We do not train models; the paper only runs inference-time evaluation. Section~\ref{sec:exp_open_book} and the experiment setup specify the evaluation protocols, metrics, the stratified ablation subset, and the ten LLMs evaluated.

\item {\bf Experiment statistical significance}
    \item[] Question: Does the paper report error bars suitably and correctly defined or other appropriate information about the statistical significance of the experiments?
    \item[] Answer: \answerNo{}
    \item[] Justification: The main results in Tables~\ref{tab:main_results} report single-pass evaluations on the full \ours{} benchmark because running every model under every protocol with multiple seeds is prohibitively expensive at this scale. Ablation experiments use three seeds and report the averaged values.

\item {\bf Experiments compute resources}
    \item[] Question: For each experiment, does the paper provide sufficient information on the computer resources (type of compute workers, memory, time of execution) needed to reproduce the experiments?
    \item[] Answer: \answerYes{}
    \item[] Justification: All evaluated models are accessed through hosted inference APIs rather than local GPUs. Closed-source models use their official providers, and open-weight models are served through Vertex AI and the official APIs of DeepSeek and DashScope, so reproduction requires only API access at the listed token volumes.
    
\item {\bf Code of ethics}
    \item[] Question: Does the research conducted in the paper conform, in every respect, with the NeurIPS Code of Ethics \url{https://neurips.cc/public/EthicsGuidelines}?
    \item[] Answer: \answerYes{}
    \item[] Justification: The authors have reviewed the NeurIPS Code of Ethics and the research conforms to it. The benchmark consists of synthetically injected agent-trajectory failures over publicly available task suites and does not collect personal data.

\item {\bf Broader impacts}
    \item[] Question: Does the paper discuss both potential positive societal impacts and negative societal impacts of the work performed?
    \item[] Answer: \answerYes{}
    \item[] Justification: Appendix~\ref{app:broader_impacts} discusses positive impacts around safer and more reliable agent deployment through systematic failure attribution, and negative impacts around the dual-use risk of red-team error injection together with our reasoning for why the benefit outweighs the marginal risk.
    
\item {\bf Safeguards}
    \item[] Question: Does the paper describe safeguards that have been put in place for responsible release of data or models that have a high risk for misuse (e.g., pre-trained language models, image generators, or scraped datasets)?
    \item[] Answer: \answerNA{}
    \item[] Justification: The released asset is a benchmark of labeled agent failure traces built from public task suites; no pre-trained model, generative system, or scraped corpus with high misuse potential is released.

\item {\bf Licenses for existing assets}
    \item[] Question: Are the creators or original owners of assets (e.g., code, data, models), used in the paper, properly credited and are the license and terms of use explicitly mentioned and properly respected?
    \item[] Answer: \answerYes{}
    \item[] Justification: \textsc{Who\&When Pro} reuses existing source benchmarks and agentic frameworks. Each upstream asset is cited at its first appearance, and we use it under the license released by its original creators.

\item {\bf New assets}
    \item[] Question: Are new assets introduced in the paper well documented and is the documentation provided alongside the assets?
    \item[] Answer: \answerYes{}
    \item[] Justification: The new asset, \textsc{Who\&When Pro}, is documented in Section~\ref{sec:trajectory-collection}--\ref{sec:stats} and in the appendix sections describing the unified error taxonomy, source datasets, and agentic frameworks.

\item {\bf Crowdsourcing and research with human subjects}
    \item[] Question: For crowdsourcing experiments and research with human subjects, does the paper include the full text of instructions given to participants and screenshots, if applicable, as well as details about compensation (if any)?
    \item[] Answer: \answerNA{}
    \item[] Justification: The human review was conducted by the authors of this paper rather than crowdsourced workers, so no external participant recruitment, instructions, or compensation arrangements apply.

\item {\bf Institutional review board (IRB) approvals or equivalent for research with human subjects}
    \item[] Question: Does the paper describe potential risks incurred by study participants, whether such risks were disclosed to the subjects, and whether Institutional Review Board (IRB) approvals (or an equivalent approval/review based on the requirements of your country or institution) were obtained?
    \item[] Answer: \answerNA{}
    \item[] Justification: The human review only inspects model-generated agent traces over public benchmark tasks. No private or personally identifying information is collected from participants or revealed in the released data, so IRB review was not required.

\item {\bf Declaration of LLM usage}
    \item[] Question: Does the paper describe the usage of LLMs if it is an important, original, or non-standard component of the core methods in this research? Note that if the LLM is used only for writing, editing, or formatting purposes and does \emph{not} impact the core methodology, scientific rigor, or originality of the research, declaration is not required.
    \item[] Answer: \answerYes{}
    \item[] Justification: We provide declaration in Appendix~\ref{app:llm}.

\end{enumerate}

\end{document}